\documentclass[10pt,journal,compsoc]{IEEEtran}

\ifCLASSOPTIONcompsoc
  \usepackage[nocompress]{cite}
\else
  \usepackage{cite}
\fi
\ifCLASSINFOpdf
  \usepackage[pdftex]{graphicx}
\else
  \usepackage[dvips]{graphicx}
\fi
\usepackage{amsmath}
\DeclareMathOperator{\sign}{sign}
\usepackage{bm}
\usepackage{caption}
\usepackage{subcaption}

\usepackage{acronym}
\usepackage{algorithmic}

\usepackage{array}

\usepackage{mdwmath}
\usepackage{mdwtab}

\usepackage{eqparbox}

\ifCLASSOPTIONcompsoc
\else
\fi

\ifCLASSOPTIONcaptionsoff
 \usepackage[nomarkers]{endfloat}
\let\MYoriglatexcaption\caption
\renewcommand{\caption}[2][\relax]{\MYoriglatexcaption[#2]{#2}}
\fi
\usepackage{url}

\ifCLASSINFOpdf
 \usepackage[pdftex]{thumbpdf}
\else
 \usepackage[dvips]{thumbpdf}
\fi

\usepackage{booktabs}

\usepackage{color}
\usepackage[dvipsnames]{xcolor}

\usepackage{lineno}

\begin{document}
%
\title{Viewpoint Selection for Photographing Architectures}
\author{Jingwu~He,
        Linbo~Wang,
        Wenzhe~Zhou,
        Hongjie~Zhang,
        Xiufen~Cui,
        and~Yanwen~Guo
\IEEEcompsocitemizethanks{\IEEEcompsocthanksitem Jingwu~He, Wenzhe~Zhou, Hongjie~Zhang and Yanwen~Guo are with the National Key Lab for Novel Software Technology, Nanjing University, NanJing, 210023, China.\protect\\
E-mail: hejw005@gmail.com, zwzmzd@mail.enjoy-what.com, hjzhang@smail.nju.edu.cn and ywguo@nju.edu.cn (Corresponding author: Yanwen Guo)
\IEEEcompsocthanksitem Linbo~Wang is with the MOE Key Laboratory of Intelligent Computing and Signal Processing, Institute of Media Computing, Anhui University, HeFei, 230039, China.\protect\\
E-mail: wanglb.2005@gmail.com
\IEEEcompsocthanksitem Xiufen~Cui is with the Samsung Electronics R\&D China.
\protect\\
E-mail: sophie.cui@samsung.com
}
\thanks{This work is supported by the National Natural Science Foundation of China under Grants 61373059, 61321491, 61672279, and 61502005, and the Natural Science Foundation of Jiangsu Province under Grant BK20150016, and the Anhui Science Foundation under Grant 1608085QF129.}
}

\IEEEtitleabstractindextext{%
\begin{abstract}
This paper studies the problem of how to choose good viewpoints for taking photographs of architectures. We achieve this by learning from professional photographs of world famous landmarks that are available on the Internet. Unlike previous efforts devoted to photo quality assessment which mainly rely on 2D image features, we show in this paper combining 2D image features extracted from images with 3D geometric features computed on the 3D models can result in more reliable evaluation of viewpoint quality. Specifically, we collect a set of photographs for each of 15 world famous architectures as well as their 3D models from the Internet. Viewpoint recovery for images is carried out through an image-model registration process, after which a newly proposed viewpoint clustering strategy is exploited to validate users' viewpoint preferences when photographing landmarks. Finally, we extract a number of 2D and 3D features for each image based on multiple visual and geometric cues and perform viewpoint recommendation by learning from both 2D and 3D features using a specifically designed SVM-2K multi-view learner, achieving superior performance over using solely 2D or 3D features. We show the effectiveness of the proposed approach through extensive experiments. The experiments also demonstrate that our system can be used to recommend viewpoints for rendering textured 3D models of buildings for the use of architectural design, in addition to viewpoint evaluation of photographs and recommendation of viewpoints for photographing architectures in practice.
\end{abstract}

}

\maketitle

\IEEEdisplaynontitleabstractindextext

%
\IEEEpeerreviewmaketitle{}

\section{Introduction}
\label{Introduction}
Modern digital cameras and smart-phones enable ordinary people to take high-quality photographs more and more easily today.
At the same time, with the rapid development of the Internet, the number of photos of architectures that can be accessed is growing explosively.
For example, people may have a holiday trip and usually want to take some photos with architectures as the background, especially when visiting those famous tourist attractions.
However, what makes for a visually-pleasing landmark photo is probably a common question raised by novice photographers.
Although people may give various answers to it, there should be no doubt that viewpoint selection plays a crucial role.
However, choosing a good viewpoint when photographing is not an easy task, especially for novices.
This quandary could be very much relieved if the photographer is provided with a useful viewpoint recommendation tool. With this in mind, we study here the relationship between viewpoint selection and the beauty of architecture photos.

There have been a number of works on the assessment of photo aesthetics concerning image
content\cite{dhar2011high}, simplicity\cite{ke2006design}, and composition\cite{luo2008photo}, but none of them are specifically designed for architecture photos. Moreover, architecture photographing is more of an experience-dependent task, and few explicit rules can be directly encoded for taking high-quality pictures. Therefore, our basic idea of setting up a viewpoint recommendation framework is to learn from photos that people rank high on the Internet. In addition, we observe that when photographing landmarks, experienced photographers also stress a lot on the sense of stereoscopic presence of the architecture. This inspires us to integrate geometric information
of the landmark model into our learning framework.

We initiate our learning framework by collecting a set of photographs for each of 15 world famous architectures as well as their 3D models from the Internet. Before stepping into the learning task, we need first recover geometric information for each image. Given a 3D architecture model and one photo, Direct Linear Transform (DLT) can be applied to a set of user-annotated corresponding points to obtain the calibration parameters of the image. However, performing viewpoint recovery for a large set of image-model pairs in this way is time-consuming.
More recently, viewpoint estimation has also been addressed by Convolutional Neural Network (CNN). Among them, Tulsiani and Malik predict the viewpoint by estimating the three Euler angles corresponding the instance with the angles predicted into several disjoint bins\cite{7298758}, and Wohlhart and Lepetit map an input image to a compact and discriminative descriptor with CNN that can be used to recognize object and estimate the 3D pose\cite{Wohlhart_2015_CVPR}. All these works require sufficient groundtruth data for the training.
In this work, we recover the 3D information of the input images in a  semi-automatic way. Specifically, we first employ Structure-from-Motion (SfM) to obtain a partial point cloud model and recover every photo's camera matrix with respect to this point cloud model. A registration process between the point cloud model and the 3D mesh model is conducted afterwards, to obtain camera matrices of the images with respect to our mesh model.
Compared with these methods, our method is much more accurate and does not need to train a learner learner beforehand.
By this way, we successfully recover the viewpoints for the entire image set without imposing considerable user burden.

To better justify the motivation behind our work on viewpoint assessment and recommendation, we further conduct studies on users' preference when photographing architectures.
To this end, we inspect the viewpoints of all the images and check whether people are accustomed to  shooting around certain locations and at specific angles given a specific architecture. This is achieved by a viewpoint clustering procedure in the underlying geometric space. Considering that the viewpoint space is essentially a Riemannian manifold with the structure of Matrix Lie Group, we introduce a Riemannian metric to measure the proximity between two viewpoints. Once the metric is defined, viewpoint preferences are successfully verified by K-medoids clustering which confirms our motivation.

Next, we focus our efforts on the task of viewpoint recommendation, which is resolved by learning from image and geometric features. Specifically, for each of the landmark photo sets we
collected, we first extract multiple image features by quantifying various visual cues and knowledge of photo aesthetics, such as histograms of oriented gradients and rule of thirds, etc.
A number of 3D features describing the 3D geometry of the architecture under a fixed viewpoint,
including project area, surface visibility, etc., are computed as well.
Both kinds of features reveal knowledge about viewpoint from different perspectives, forming multi-modular information for viewpoint estimation. Learning from the features can be conducted by training single-view learners from 2D and 3D features separately and fusing them together finally. This way, though simple, is however difficult to harness their mutual knowledge.  Instead, we choose to exploit a multi-view learning framework, whose paradigm is specifically defined for the task of multi-modular data analysis.

Multi-view learning has been widely studied so far. Among all the techniques developed, we choose SVM-2K\cite{farquhar2005two} as our viewpoint learner.
SVM-2K combines the two stage learning (KCCA followed by SVM) into a single pass of optimization, and it is specifically designed to deal with two types of features.
To start learning, we further conduct an user study by asking the participants to rank the goodness of training photos.
With all these prepared, we perform training with multi-view learning of SVM-2K.
In addition, learning results on solely 2D or 3D features are also reported. It shows less comparative performance than using both kinds of features simultaneously, verifying the necessity of considering both image and geometric knowledges when photographing. Besides, classification performance of each individual feature is also validated, which provides users with valuable rules that should be paid attention to when photographing architectures.

In summary, the main contributions of this paper are three-fold.
\begin{itemize}
\item We propose a new framework for viewpoint analysis as well as recommendation
when photographing architectures. Both image and geometric features are fed into SVM-2K, a multi-view learner, to learn good viewpoints, achieving considerable performance improvement over single-view learning.
\item User preference over viewpoints is analyzed by a clustering process with an effective distance measure newly introduced to describe the proximity of viewpoints. Results suggest people tend to have consistent viewpoint preference when photographing architectures, justifying the necessity of viewpoint recommendation.

\item Our system can be used to evaluate the goodness of viewpoints of photographs and to recommend viewpoints for photographing architectures, even in the absence of geometric models, benefited from the multi-view learner. In addition, our system can also be used to recommend viewpoints for rendering textured 3D models of buildings for the use of architectural design.

\end{itemize}


A preliminary, short version of this paper was published as a conference version~\cite{pg.20161333}. Compared with that version, this paper first reveals more technical details, such as viewpoint recovery for a large set of images and the definition of features. Furthermore, we now use SVM-2K for learning from both image and geometric features and confirm that using both kinds of features shows superior performance over learning with only either aspects of them. With extensive experiments, we show more applications of our framework as well.


\section{Related work}




\subsection{Photo aesthetic assessment}
Aesthetic assessment of photographs has been investigated by previous methods. Many researchers devote their efforts to learning from visual features, including image composition\cite{luo2008photo,Datta2006,guo2012improving}, content\cite{dhar2011high,6126498}, and simplicity\cite{ke2006design}, etc. Generic image features such as Bag-of-visual-words, Fisher Vector \cite{marchesotti2011assessing} are also used to assess the aesthetic quality of photographs. Datta et al. proposed a computational approach to understand what aspects of a photograph appealed to people from a population and statistical standpoint \cite{Datta2006}. High-level describable attributes of images that are useful for predicting perceived aesthetic quality of images are also be estimated \cite{dhar2011high}. Luo et al. proposed a content-based photo quality assessment framework with both regional and global features \cite{6126498}. Furthermore, assessment of aesthetic quality is also addressed by deep learning\cite{Chang:2016:ATP:2897824.2925908}. Tian et al. mined the underlying aesthetic attributes automatically with deep convolutional neural networks (DCNNs)\cite{7271097}.


Besides, there are two important criteria often used by professional photographers to evaluate photo quality: low-level features and composition. Low-level features include the exposure, contrast, colorfulness, and textures. Composition means the organization of all the graphic elements inside a photo\cite{guo2012improving}. Typically, low-level features and composition are jointly used to perform the analysis of how to make a specific image appreciated by most viewers\cite{marchesotti2011assessing}. Hence, photographs taken by experienced photographers adhere to several rules of composition, which make them more visually appealing than those taken by amateurs\cite{bhattacharya2010framework}.

The methods above mainly resort to visual features of 2D images to assess the quality of a photo. However, the sense of stereoscopic presence is also an essential component of photo assessment, which is closely related to viewpoint selection for taking a good photograph.

\subsection{General viewpoint analysis}





Viewpoint analysis is previously addressed in many different fields. Several works are designed to manipulate the images with user interaction\cite{Guo2014174,Carroll:2010:IWA:1833349.1778864,Zheng:2012:IIC:2185520.2185595}, which can be used to generate some new viewpoints of the scene. Guo et al. proposed an efficient view manipulation method for cuboid-structured images that enables the generation of novel images with new viewpoints given only a single image as input, with moderate user assistance\cite{Guo2014174}. Carroll et al. proposed an interface that allows users to manipulate perspective in photographs and it can also plausibly simulate moderate changes in perspectives induced by the controls, e.g., vanishing points \cite{Carroll:2010:IWA:1833349.1778864}.

Besides, a number of works study the problem of viewpoint selection for 3D models~\cite{3DOR:3DOR10:015-022,mortara2009semantics} and are extended to various applications~\cite{liu2012web,snavely2008finding}. Moreover, Bae et al. proposed computational rephotography that allows photographers to take modern photos matching historical photographs~\cite{bae2010computational}. An extensive review of view selection in computer graphics is available in Camera Control\cite{christie2008camera}.
Several geometric descriptors such as view entropy\cite{vazquez2003automatic} and mesh saliency\cite{lee2005mesh}, assume that the best view of an object is the one that maximizes the value of the corresponding descriptors. Laga proposed a framework for the automatic selection of the best views of 3D models, which defines the best views of a 3D object as the views that allow to discriminate the object from the other objects in the database\cite{3DOR:3DOR10:015-022}.
The semantics of the displayed features should also be related to the quality of a view, and the best view should be evaluated taking into consideration the meaningful components\cite{mortara2009semantics}.
Secord et al. leveraged the results of a large user study to optimize the parameters of a general model for viewpoint goodness, such that the fitted model can predict people's preferred views for a broad range of objects\cite{secord2011perceptual}.

Furthermore, to automatically select the best views of a 3D shape, Liu et al. proposed a web-image driven approach to reflect human perception\cite{liu2012web}. With lots of photos taken from a variety of viewpoints, approaches for browsing the photos\cite{snavely2006photo} and finding paths\cite{snavely2008finding} for the famous scene are raised. In addition, based on the approaches, users can select the viewpoints for architecture interactively. Leifman et al. took both local and global distinctness into consideration for surface regions of interest for viewpoint selection\cite{6247703}.

\subsection{Multi-view learning}
Unlike previous works, we argue in this paper that combing the 2D image features and geometric features can result in more reliable evaluation of viewpoint quality, and hence can better facilitate various applications.
In recent years, lots of learning methods from multi-view data by considering the diversity of views have been proposed. By complementing properties of different views, multi-view learning performs more effectively, more promisingly, and has better generalization ability than single-view learning\cite{xu2013survey}.
Canonical Correlation Analysis (CCA)\cite{hotelling1936relations} as well as Kernel Canonical Correlation Analysis (KCCA)\cite{0609071} is a popular and successful multi-view learning approach for mapping two views into the same latent space where correlation between the two views is maximized.
In many cases, the data can be described by more than one view, and the multi-view learning algorithm can help us considering the diversity of different views. With the knowledge of this, we conduct our viewpoint recommendation with respect to multi-view learning of SVM-2K~\cite{farquhar2005two}.

\section{Overview}
\label{Overview}
Given a set of photographs and the corresponding 3D models, our goal is to perform viewpoint assessment for the photographs. We also aim to recommend viewpoints for photographing architectures when the user is visiting famous tourist attractions and to suggest optimal viewpoints for rendering textured 3D models of buildings for the use of architectural design.

\begin{figure*}[tb]
\centering 
\includegraphics[width=\linewidth]{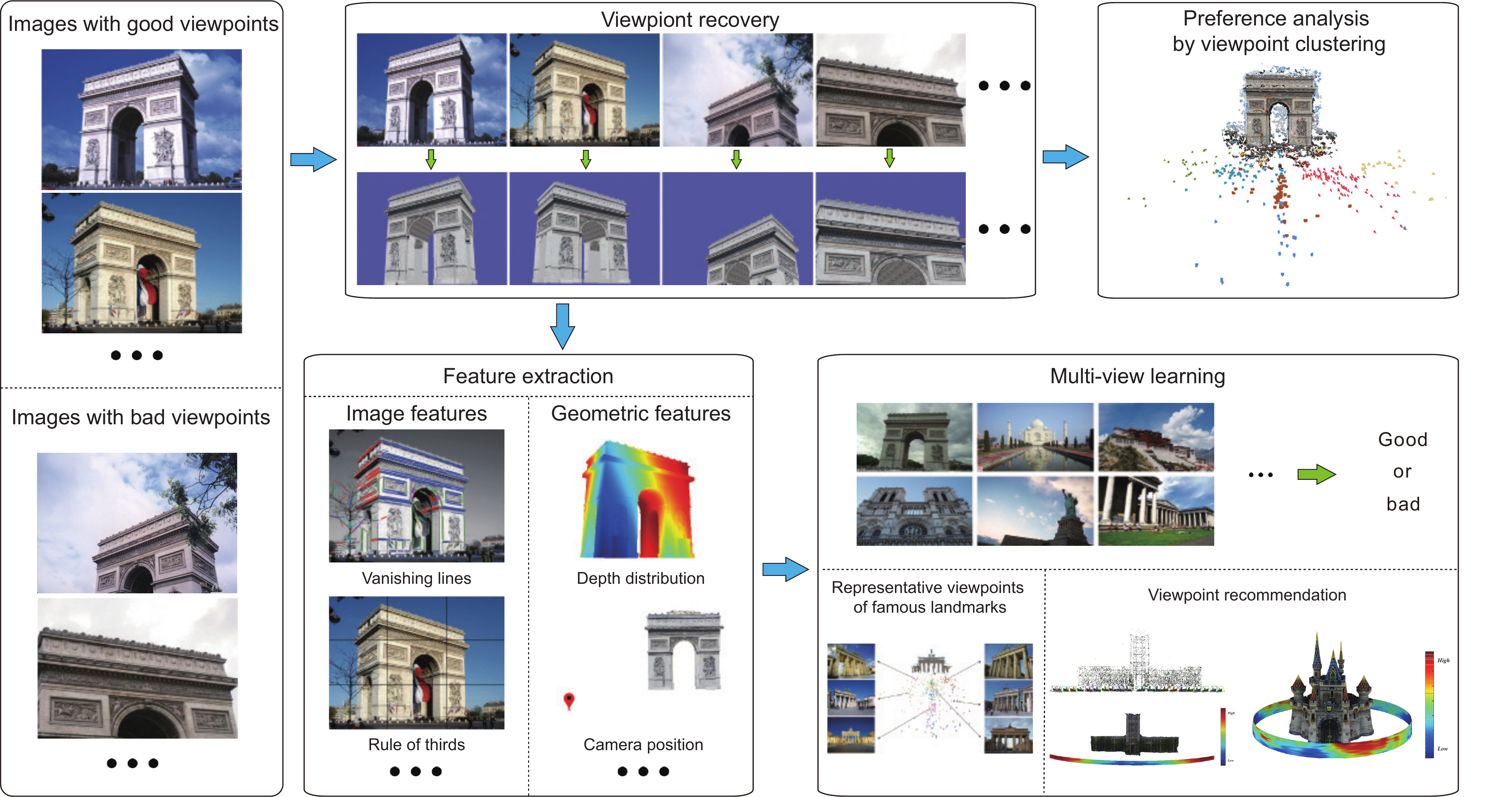}
\caption{The systematic overview of our viewpoint learning framework.}	
\label{fig:flowchart}	
\end{figure*}

A schematic overview of the proposed system is illustrated in Fig.~\ref{fig:flowchart}.
Specifically, we have collected a set of photographs for each of 15 world famous landmarks and obtained their corresponding 3D models. With these at hand, we first estimate the viewpoint for each input image, followed by a viewpoint clustering procedure to verify viewpoint preferences when photographing these architectures (Sec.~\ref{Viewpoint Estimation}). Afterwards, we extract a number of 2D and 3D features for each image based on multiple image and geometric cues and perform viewpoint assessment by feeding all the features into SVM-2K, a multi-view learning method. Viewpoint recommendation can be conveniently made by the learned classifier through selecting the viewpoint maximizing the viewpoint preference value (Sec.~\ref{Learning good views}). We finally conduct extensive experiments to show the effectiveness and wide applications of our framework (Sec.~\ref{Applications_Experiments}). In this Section, we also show the performance obtained with using only each individual image feature or 3D geometric feature, in order to provide users with simple guidance for shooting architectures.




\section{Viewpoint Preference Analysis}
\label{Viewpoint Estimation}

Given a set of images and the 3D mesh model they depict, we seek to recover the viewpoint of each image with regard to the mesh model and investigate users' viewpoint preference of photographing. In the former step, we first employ the classical
SfM~\cite{brown2005unsupervised}
to construct a point cloud model and register it with the provided 3D mesh model in an interactive manner. For each image, its viewpoint with respect to the mesh model is obtained by viewpoint propagation from the reconstructed model. In the latter step, viewpoints of all images are gathered together to form a viewpoint space, in which we analyze user preference of architecture photographing via viewpoint clustering with a newly defined distance metric. We detail the process below.

\subsection{Model registration}
\label{Model Registration}

    SfM is a classical tool used to reconstruct the 3D model from multiple images. We refer the readers to~\cite{brown2005unsupervised} for more details. After applying SfM to the input photo set, we obtain a coarse point cloud model as well as the camera matrix of each photo with respect to the model, encoding the viewpoint information. A model registration process is then conducted between the reconstructed coarse model and the provided fine model for viewpoint propagation afterwards.

    In general, the point cloud model can be aligned with the 3D mesh model after applying an affine transformation to it, adapting to the difference in scale, rotation, and translation. Therefore, the key task here is to derive the affine transformation between the two models. To this end, we first manually annotate multiple corresponding points on the two models. Let $(\mathbf{p}_{k}, \mathbf{q}_{k})_{k=1,2,...,n}$ denote $n$ pairs of matched points on the mesh model and point cloud model, respectively, and $c$, $\mathbf{R}$, and $\mathbf{t}$ represent the scale, rotation, and translation parameters encoded in the affine transformation. We have,

    \begin{equation}
        c\mathbf{R}\mathbf{p}_{k} + \mathbf{t} = \mathbf{q}_{k}.
    \end{equation}

\begin{figure}[tb]
    \centering
    \begin{subfigure}[b]{0.15\textwidth}
    \includegraphics[width=\textwidth]{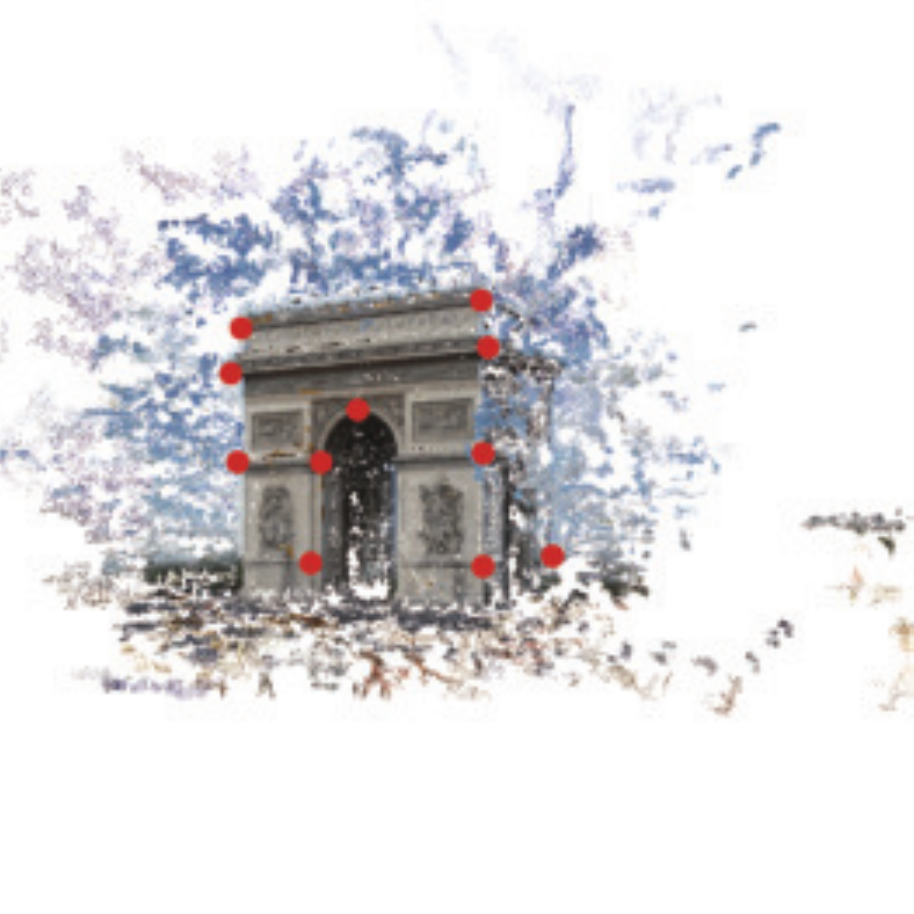}
    \end{subfigure}
    \begin{subfigure}[b]{0.15\textwidth}
    \includegraphics[width=\textwidth]{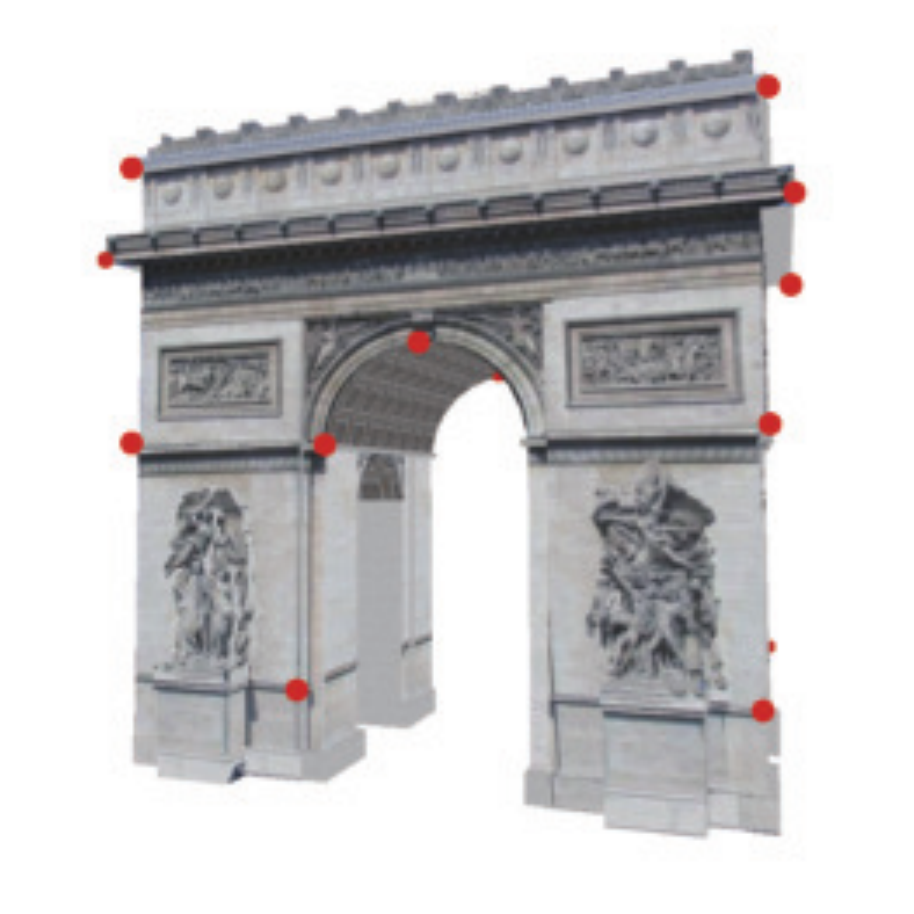}
    \end{subfigure}
    \begin{subfigure}[b]{0.15\textwidth}
    \includegraphics[width=\textwidth]{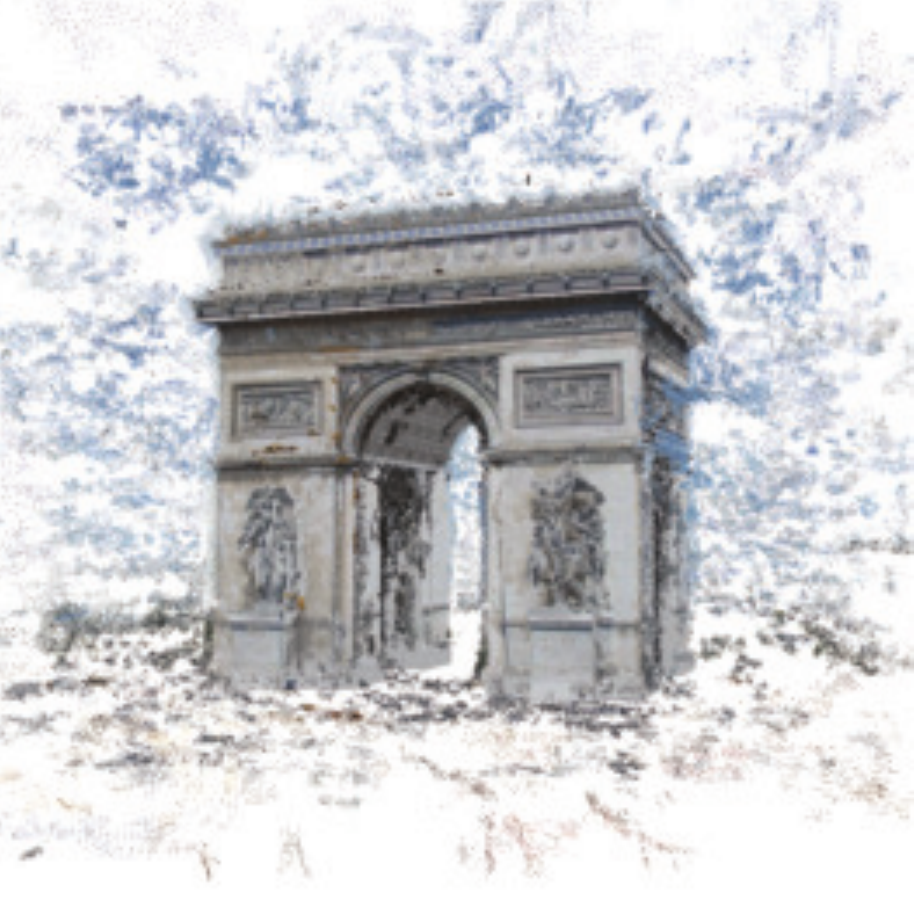}
    \end{subfigure}
    \begin{subfigure}[b]{0.15\textwidth}
    \includegraphics[width=\textwidth]{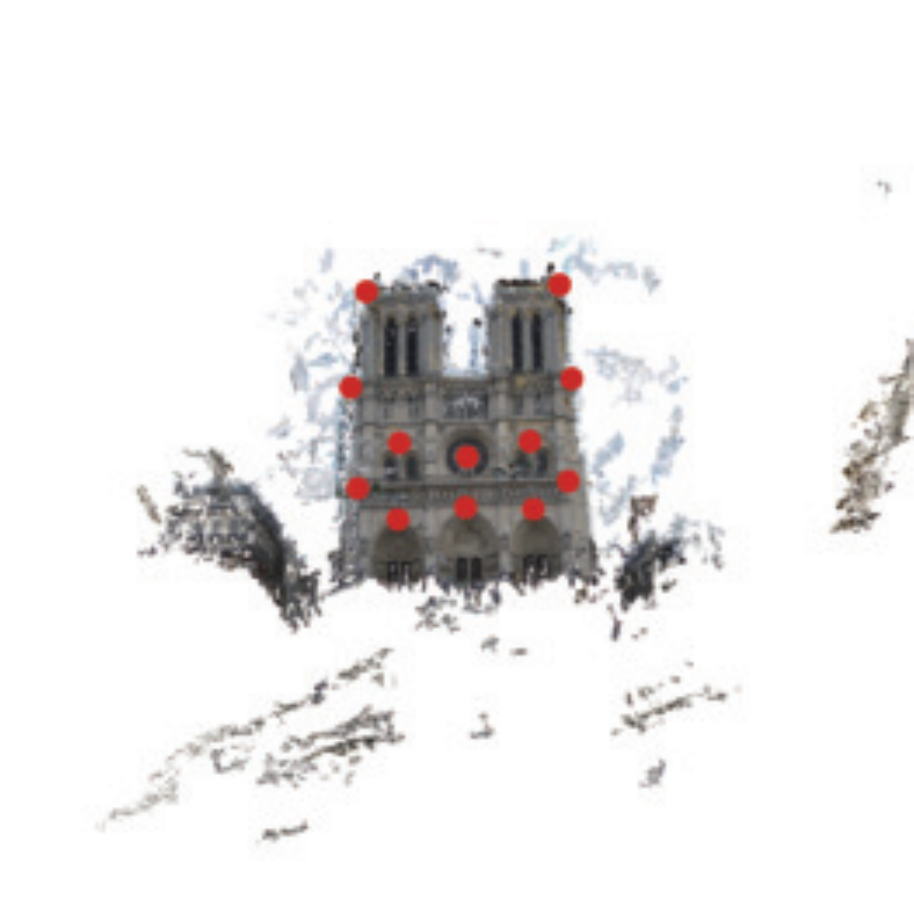}
    \end{subfigure}
    \begin{subfigure}[b]{0.15\textwidth}
    \includegraphics[width=\textwidth]{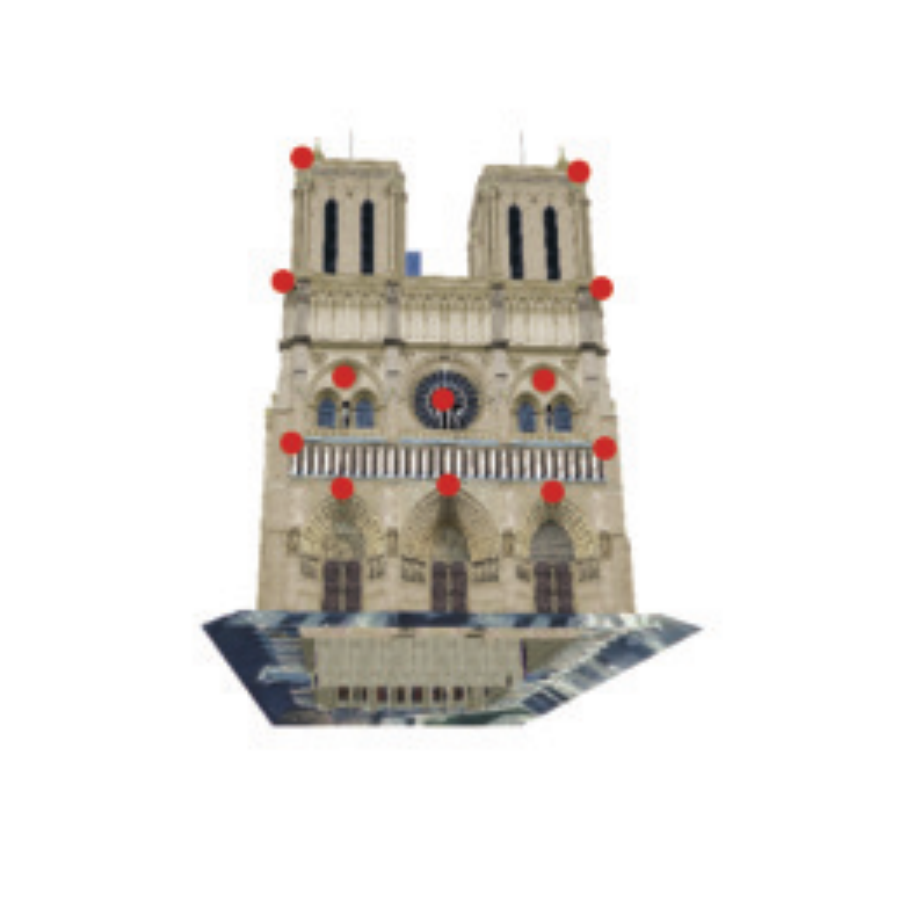}
    \end{subfigure}
    \begin{subfigure}[b]{0.15\textwidth}
    \includegraphics[width=\textwidth]{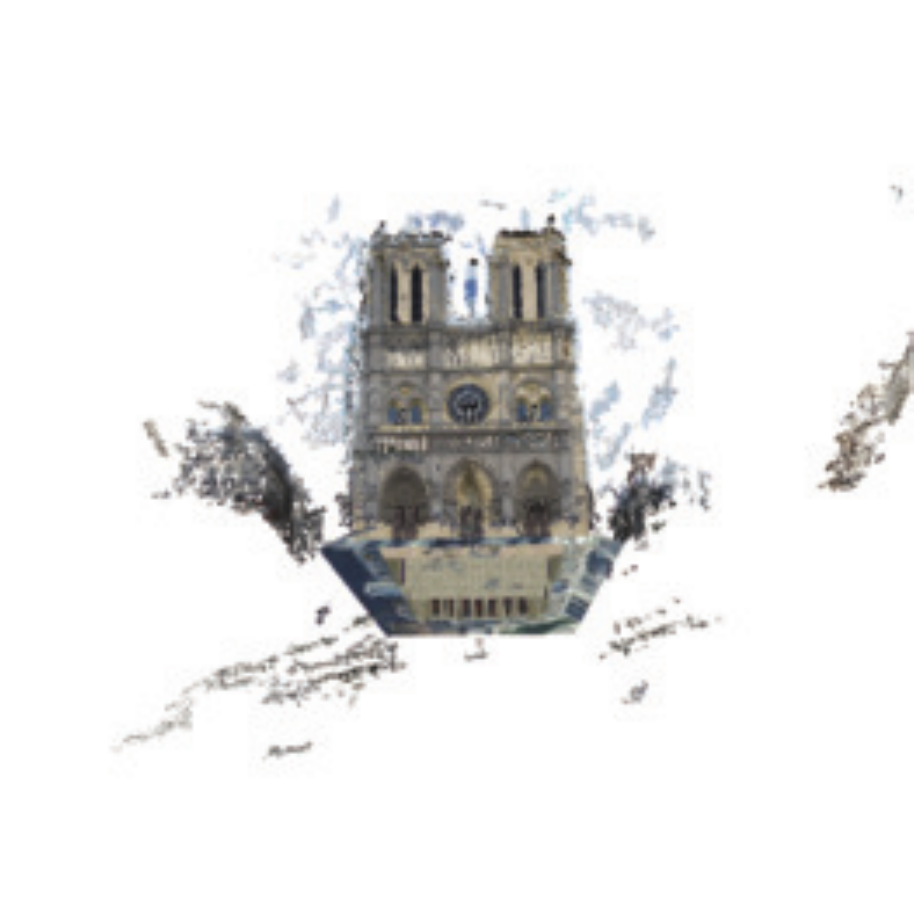}
    \end{subfigure}
    \caption{Registration between the point cloud model (left) and mesh model (middle) based on the marked points (red). We align the mesh model with the point cloud model using the registration parameters. The transformed mesh model coincides with the point cloud model (right).}
    \label{fig:model Adjustment}
    \end{figure}

    After aggregating all the $n$ equations, the parameters $c$, $\mathbf{R}$, and $\mathbf{t}$ can be obtained by solving the linear equation system. Alternatively, considering the over-constrained nature of the linear system and annotation error, it is better to solve the following energy function:
    \begin{equation}
    \label{eq:registrationEqu}
        \min E(c,\mathbf{R}, \mathbf{t}) = \sum_{k=1}^{n}\|\mathbf{q}_{k} - (c\mathbf{R}\mathbf{p}_{k} + \mathbf{t})\|.
    \end{equation}
    We use the Levenberg-Marquardt algorithm~\cite{levenberg1944method} in the LEVMAR package~\cite{lourakis04LM} to solve the optimization problem. In Fig.~\ref{fig:model Adjustment}, two examples are shown to illustrate the registration process.

    To this end, the affine transformation is estimated.

\subsection{Transformation transferring}
\label{Transformation Transferring}

To this end, we have estimated the affine transformation mapping the provided mesh model to the reconstructed point cloud model as well as the camera matrix of each image regarding to the point cloud model. Next, we obtain the viewpoint information of each image regarding to the mesh model. More specifically, supposing $\mathbf{p}$ and $\mathbf{q}$ are two corresponding points on the mesh model and the point cloud model, respectively. We have,
    \begin{equation}
        \mathbf{q} = c\mathbf{R}\mathbf{p} + \mathbf{t},
    \end{equation}
where $c$, $\mathbf{R}$, and $\mathbf{t}$ are the scale, rotation and translation parameters optimized in Eq.~\eqref{eq:registrationEqu}. Now, assume that $\mathbf{m}$ is a point in the camera coordinate system with its corresponding point on the 3D model as $\mathbf{p}$.
Meanwhile, let $\mathbf{R'}$ and $\mathbf{t'}$ denote the external camera parameters of image $I$ with regard to the point cloud model. We then have,
\begin{equation}
\label{eq:Transformation transferring}
\begin{aligned}
    \mathbf{m} & = \mathbf{R'}\mathbf{q} + \mathbf{t'} \\
           & = \mathbf{R'}(c\mathbf{R}\mathbf{q} + \mathbf{t}) + \mathbf{t'} \\
           & = c\mathbf{R'}\mathbf{R}\mathbf{q} + (\mathbf{R'}\mathbf{t} + \mathbf{t'}).
\end{aligned}
\end{equation}
Obviously, the above equation shows the coordinate transformation from a point $\mathbf{p}$ in the coordinate system of a 3D mesh model to a point $\mathbf{m}$ in the camera coordinate system,
with $c\mathbf{R'}\mathbf{R}$ and $\mathbf{R'}\mathbf{t} + \mathbf{t'}$ being the rotation and translation parameters of $I$ with respect to the mesh model respectively.

\begin{figure}[tb]
\centering
\includegraphics[width=\linewidth]{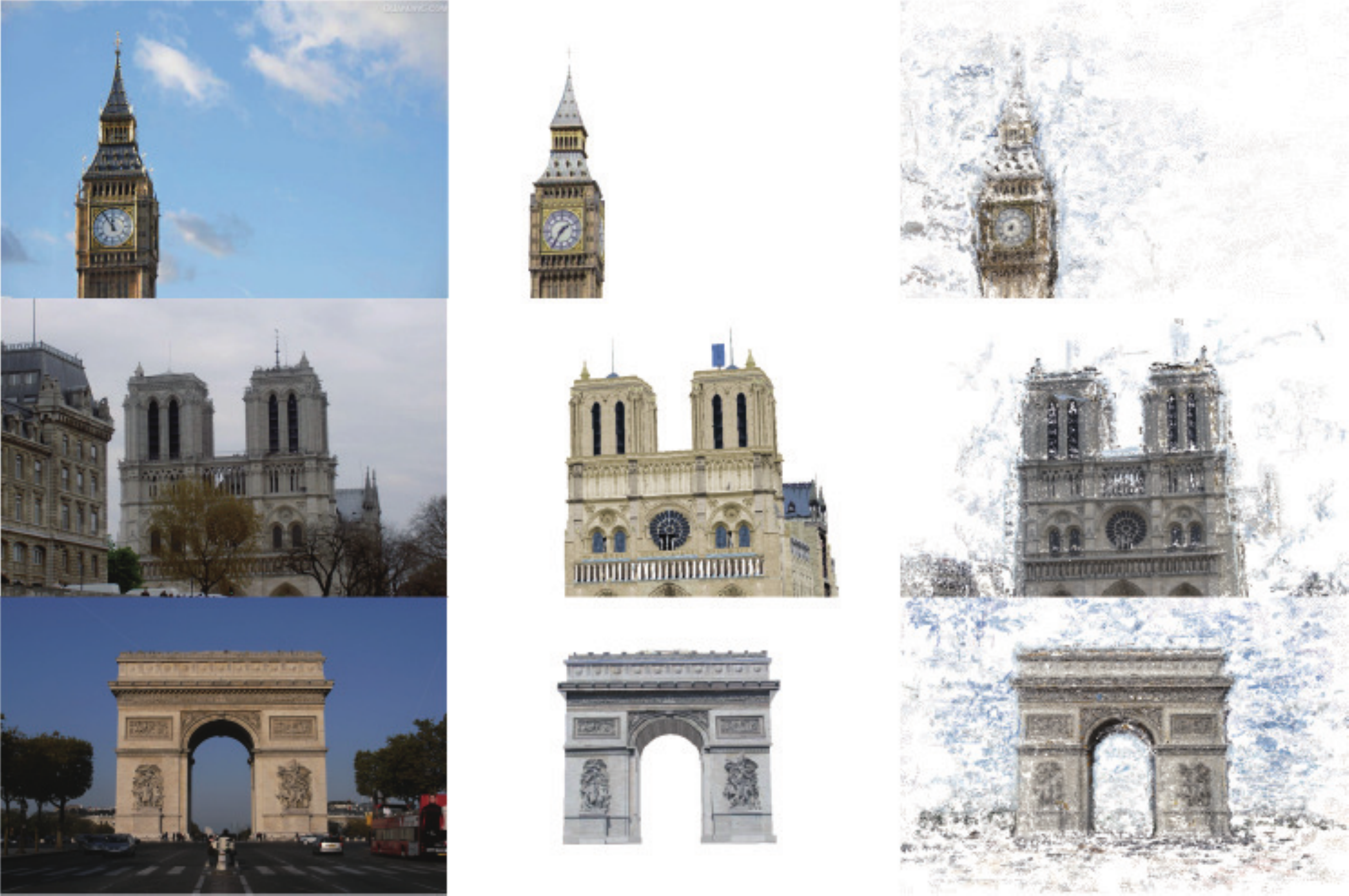}\\
\caption{Rendering the mesh model (middle) and the point cloud model (right) under the recovered viewpoint shows that our estimated viewpoint is accurate.}
\label{fig:pic pt mesh comparison}
\end{figure}

Note that Eq.~\eqref{eq:Transformation transferring} leads to an informal rotation if c does not equal to $1$. Fortunately, since the point $\mathbf{m}$ and $\frac{\mathbf{m}}{c}$ are mapped to the same point after perspective division, we use $\mathbf{R'}\mathbf{R}$ and $\frac{\mathbf{R'}\mathbf{t} + \mathbf{t'}}{c}$ as the final external camera matrix parameters of the image $I$. In Fig.~\ref{fig:pic pt mesh comparison}, we illustrate the viewpoint recovery results by showing comparative images of three models. For each model, one is the original image and the other two are images generated by rendering the given mesh model and the point cloud model, respectively.



\subsection{Viewpoint clustering for preference analysis}

\label{subsec_view_clustering}


Given a set of photos, we estimate their viewpoints regarding to the provided 3D model as described above.
Now, we study whether people tend to shoot at some specific angles when standing in front of an architecture.
We achieve this by performing a clustering procedure in the underlying geometric space.

A photo of an architecture may be regarded as a visual experience by looking at the model of the architecture
from a specific angle and holding on at a certain point. More formally, it is to project
the model of the architecture with a model-view matrix and a projection matrix.
As the two matrices together determine the content of a photo, the model-view matrix
records various parameters about camera, including position, orientation, etc.,
and plausibly dominate the viewers' sense of viewpoint for a model. The projection matrix,
on the other hand, contains information about the projection plane, imposing more
influence on the size of the architecture presented in the photo. Therefore, we
focus on model-view matrices recovered from the given photo set for
analysis of viewpoint preference.

We stack up all the model-view matrices together to form a viewpoint space. Note that
this is in essential not a vector space, in which Euclidean distance metric is embedded.
Moreover, it is one kind of matrix Lie groups, equipped with the structure of analytic
Riemannian manifold. Unlike existing approaches relying on heuristic metric, e.g. homography overlap
distance~\cite{weyand2011discovering, weyand2013discovering}, to describe the viewpoint
similarity among photos, we introduce a Riemannian metric that specifically designed for
matrix Lie groups \cite{rossmann2002lie,7274720} to define the distance between two model-view matrices,
and thus better measures the viewpoint proximity of two photos. More formally,
given two model-view matrices $\textbf{M}_i$ and $\textbf{M}_j$, it is defined as
\begin{equation}
\label{eq:distance metric}
\begin{aligned}
  d(\textbf{M}_i,\textbf{M}_j)={\|\log({\textbf{M}_i}^{-1}{\textbf{M}_j})\|}_F,
\end{aligned}
\end{equation}
where ${\|.\|}_F$ denotes the Frobenius norm of a matrix.

\begin{figure}[tb]
\centering
    \includegraphics[width=\linewidth]{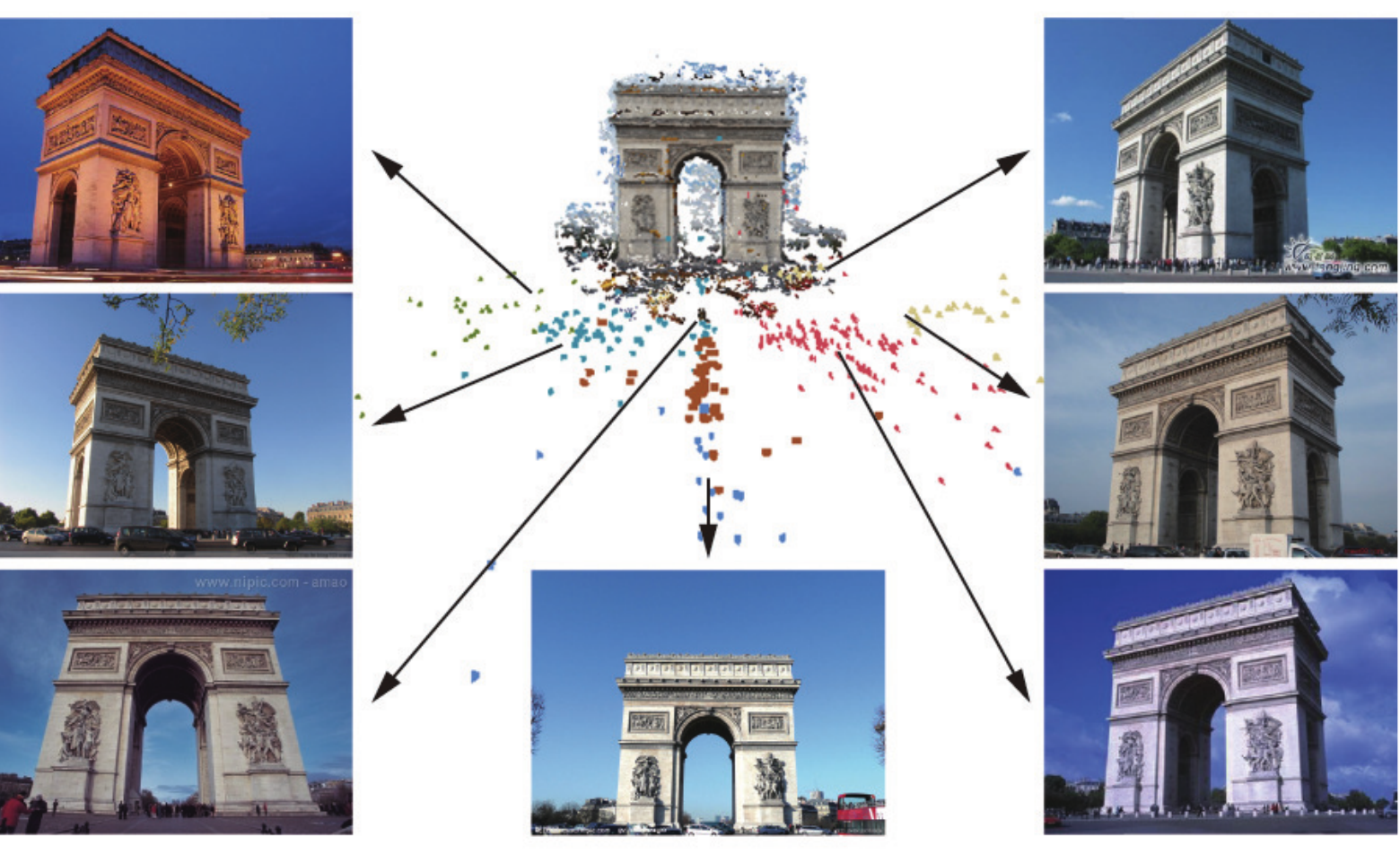}
    \caption{Result of viewpoint clustering with seven representative viewpoints shown here.}
    \label{fig:kmedoidsClustering}
\end{figure}

With distance metric defined above, we perform K-medoids clustering for viewpoint preference analysis.
More sophisticated clustering algorithm, such as Mean-shift clustering, can be applied here.
We choose the classical K-medoids algorithm with $K=9$, considering its simplicity and being easy to implement.
Moreover, it suffices to show that people often share consistent viewpoint preferences
when photographing architecture in our experiments,
as illustrated by Fig.~\ref{fig:kmedoidsClustering}.
This is in compliance with our common knowledge and justifies the necessity for viewpoint recommendation.

\section{Learning good views}
\label{Learning good views}

Our main observation of the proposed viewpoint learning method is that professional photographers usually rely on various visual and geometric cues embodied in the scene to choose the optimal viewpoints of photographing. Therefore, we first extract multiple 2D image features as well as 3D geometric features from the image and the corresponding model, separately. Thereafter, we choose a multi-view learner, namely SVM-2K\cite{farquhar2005two} which accepts the 2D image features and 3D geometric features for training, in order to explore the intrinsic relationship between the features and good viewpoints. We also demonstrate its superior performance over the learner based on solely 2D image or 3D geometric features.



\subsection{Feature extraction}
\label{Feature extraction}
We introduce the 2D image and 3D geometric features extracted for the learning task in this subsection.


\subsubsection{Image features}
Taking high quality photographs of an architecture needs to consider many aspects, such as color harmony, shape of the architecture, composition of the photo, and so forth.
We extract the following low-level visual and high-level semantic features from the image.

\begin{itemize}

\item $\mathbf{v}_{color}$: \emph{Color entropy and distribution.}
Both features are defined by making statistics about pixels' RGB values in the image. We quantize each of the red, green, and blue channels into 8 bins. Then we can get a histogram with $512$ bins. With this histogram, we compute color entropy and distribution as,
\begin{equation}
\begin{aligned}
c_e &= \sum P_i \log P_i, \\
c_d &= 1 - \sum P_i^2, &~~\text{with } P_i = \frac{\mathbf{H}(i)}{\sum \mathbf{H}(i)},
\end{aligned}
\end{equation}
where $\mathbf{H}$ represents the histogram, and $c_e$ and $c_d$ denote color entropy and distribution, respectively.
We then have $\mathbf{v}_{color} = [c_e,~c_d,~r_m,~g_m,~b_m]$ with $r_m$, $g_m$, and $b_m$ being the mean values of RGB channels.


\item $\mathbf{v}_{bright},\mathbf{v}_{contrast}$: \emph{Brightness and contrast.} Brightness and contrast are low level features~\cite{ke2006design}. They could be very different under different viewpoints.

\item $\mathbf{v}_{blur}$: \emph{Blur.}
Image sharpness is arguably one of the most important factors influencing image quality.
To measure image sharpness, we compute the blur degree of the input image by using the metric proposed by~\cite{ke2006design}.

\item $\mathbf{v}_{hsv}$: \emph{Hue count, hue histogram, hue entropy, Saturation histogram, and Saturation entropy.} The hue count $h_c$ of a photo is a measure of its simplicity. \cite{ke2006design}. The smaller the $h_c$ value is, the more colorful the photo looks.
We define
$\mathbf{v}_{hsv} = [h_c,~h_h,~h_e,~s_h,~s_e]$ where $h_c$ is the hue count, and $h_h$, $h_e$, $s_h$, and $s_e$ are histogram and entropy of hue and saturation in HSV color space, respectively.

\item $\mathbf{v}_{HOG}$: \emph{HOG (Histograms of oriented gradient).} HOG is defined as the concatenation of local histograms of gradient directions. The spirit behind is that local object appearance and shape within an image can be described by the distribution of edge directions. The feature varies under different viewpoints.

\item $\mathbf{v}_{vl}$: \emph{Vanishing lines.} Vanishing lines are important visual features, especially for the architecture photos we focus on. We detect three vanishing lines $[l_a,~l_b,~l_c]$ which correspond to the three dominant vanishing points with the method proposed in \cite{Li2010}.
We define $\mathbf{v}_{vl}$ as the set of three angles each of which is the angle between every two vanishing lines.

\item $\mathbf{v}_{pc}$: \emph{Photo composition.}
Image composition serves as a crucial high-level aspect influencing visual aesthetics. A simple, yet intuitive guideline is the rule of thirds which means that an image should be imaged as divided into nine equal parts by two equally-spaced horizontal lines and two equally-spaced vertical lines, and important compositional elements should be placed along these lines or their intersections. We compute this feature by using the metric given in~\cite{guo2012improving} by taking the architecture as the foregound object.

\end{itemize}

\subsubsection{Geometric features}

A few methods are developed for selecting good viewpoints given a mesh model by considering the geometric cues such as view entropy\cite{vazquez2003automatic} and mesh saliency\cite{lee2005mesh}. Following them, we extract the following frequently used geometric features as well as some intuitive features defined according to empirical rules that are commonly used for photographing architectures.


\begin{itemize}

\item $\mathbf{g}_{mc}~and~\mathbf{g}_{gc}$: \emph{Mean curvature and gaussian curvature.} These two features are the most commonly used features in geometric processing.

\item $\mathbf{g}_{md}~and~\mathbf{g}_{dd}$: \emph{Max depth and depth distribution.} Depth features are useful to help avoid degenerated viewpoints. $\mathbf{g}_{dm}$ is defined as the maximum depth value of visible points on the shape. Depth distribution is introduced to encourage a broad, even distribution of depths in the scene\cite{secord2011perceptual}.

\item $\mathbf{g}_{area},\mathbf{g}_{surface},\mathbf{g}_{ve}$: \emph{Project area, surface visibility, and viewpoint entropy.}
Project area is defined as the ratio between the projected area of the model and the image size.
Surface visibility depicts the amount of hidden surface of an object \cite{secord2011perceptual}.
Viewpoint entropy quantifies the amount of information that can be captured from a specific viewpoint \cite{vazquez2001viewpoint}.
These three features all relate to the shape area as seen from a particular viewpoint.

\item $\mathbf{g}_{outer}$: \emph{Outer points.} We always want to take a photo including the whole architecture. We thus set $\mathbf{g}_{outer}$ to the ratio of mesh points out of the photo to all points on 3D model.
\begin{equation}
\begin{aligned}
    \mathbf{g}_{outer} = \frac{\sum_{i \in \text{outer}}}{\sum_{i \in \text{all}}}.
\end{aligned}
\end{equation}

\item $\mathbf{g}_{sl},\mathbf{g}_{sc}$, and $\mathbf{g}_{sce}$: \emph{The length, curvature, and curvature extrema of the projected model silhouette.} Silhouette attributes are believed to be the first index into the human memory of shapes. Silhouette length defines the overall length of the object silhouette in the image plane. Silhouette curvature is introduced as a visual feature. It provides significant information to the viewer\cite{feldman2005contours,vieira2009learning}.

\item $\mathbf{g}_{pos}$: \emph{Camera position.} We take the spherical coordinate $[r,~\theta,~\phi]$ as the camera position. $\mathbf{g}_{pos} = [\theta,~\phi]$ is used as the position feature for a given viewpoint.

\item $\mathbf{g}_{ut}$: \emph{Up-direction tilt.} This feature is defined as the cosine angle between the up direction $\mathbf{uc}$ of the camera system and the up direction $\mathbf{ua}$ of the architecture.
\begin{equation}
\begin{aligned}
\mathbf{g}_{ut} = \cos(\mathbf{uc},\mathbf{ua}).
\end{aligned}
\end{equation}

\item $\mathbf{g}_{angles}$: \emph{Axis angles.}
Let $\mathbf{x}_m$, $\mathbf{y}_m$, $\mathbf{z}_m$ and $\mathbf{x}_c$, $\mathbf{y}_c$, $\mathbf{z}_c$ be the axes of the world coordinate system and the camera coordinate system, separately. We have,

\begin{equation}
\begin{aligned}
	\mathbf{g}_{angles} = \angle(\mathbf{u}_{m},\mathbf{u}_{c}),
\end{aligned}
\end{equation}
where $\mathbf{u}_{m}\in\{\mathbf{x}_m,~\mathbf{y}_m,~\mathbf{z}_m\}$ and $\mathbf{u}_{c}\in\{\mathbf{x}_c,~\mathbf{y}_c,~\mathbf{z}_c\}$.



\item $\mathbf{g}_{ap}$: \emph{Above preference.} People tend to prefer views that are slightly above the horizon\cite{blanz1999object}. We define this feature to evaluate the preference of viewpoint.
\begin{equation}
\begin{aligned}
\mathbf{g}_{ap} = \mathcal{G}(\phi; \frac{3\pi}{8}, \frac{\pi}{4}),
\end{aligned}
\end{equation}
where $\phi$ is the latitude of the viewpoint position on the viewing sphere, and $\mathcal{G}$ is a Gaussian function.

\end{itemize}




\subsection{Multi-view learning for viewpoint selection}
\label{multi view learning}

As described above, in our applications, features are extracted in two views, namely, 2D image features and 3D geometric features, both containing common knowledge about viewpoint. Features in each view could be used to build a viewpoint model for predicting the viewpoint quality. However, the features extracted also contain some distracting information, e.g., the 2D image features are generally extracted by quantizing various visual cues regarding to image quality, which can be impacted by many factors except for viewpoint. To harness the mutual knowledge between the two-view features as well as to be better immune to the distracting noisy information in each single view during learning, we choose to train a multi-view learner for the learning task rather than the single-view one.

Among all the multi-view learning algorithms, we adopt SVM-2K as our learner. It combines KCCA with SVM, and is originally proposed for two-view classification. In contrast to the standard KCCA aiming to achieve correlation maximization between two-view feature projections, it exploits a distance minimization version of KCCA. More specifically, in our settings, we use SVM-2K to train two SVM based viewpoint quality learners simultaneously, with one in the image feature space and the other in the geometric feature space. Additional constraints are imposed to minimize the disagreement between the image and geometric viewpoint models and to force the outputs of the two SVMs (alternatively, the projections of the image and geometric feature vectors of the same photo on the weight vectors of the two SVMs) as close as possible. Since viewpoint quality forms the common knowledge residing in the two-view feature space, the joint learning process leads to a reasonable amount of performance improvement over the single-view learner.
Next, we introduce the SVM-2K based learner for viewpoint quality learning.


Let us denote $\mathbf{x}_i$ as the feature vector of the input image $\emph{I}_i~(i=1, 2,..., n)$ with $\mathbf{x}_i$ formed by all the image and geometric features extracted in Sec.~\ref{Feature extraction}. $\phi_V(\mathbf{x}_i)$ is a kernelized feature vector of the image features in $\mathbf{x}_i$ using the kernel $\kappa_V$, while $\phi_G(\mathbf{x}_i)$ is similarly defined for geometric features in $\mathbf{x}_i$ using the kernel $\kappa_G$. We use SVM-2K to learn two SVM classifiers $(\mathbf{w}_V, b_V)$ and $(\mathbf{w}_G, b_G)$ with the standard SVM formulation from the training data $\{(\mathbf{\phi}_V(\mathbf{x}_i), \mathbf{\phi}_G(\mathbf{x}_i), y_i)~|~i = 1,2,...,n\}$, where $y_i \in \{-1, 1\}$ is the label of viewpoint quality of image $\emph{I}_i$. The disagreement between the two linear functions with regarding to the two SVM learners is minimized by solving the following energy function,
\begin{small}
\begin{align}
	\min_{\substack{\mathbf{w}_V,b_V,\\\mathbf{w}_G,b_G}}&\frac{1}{2}{\|\mathbf{w}_V\|}^2+\frac{1}{2}{\|\mathbf{w}_G\|}^2+C^V\sum_{i=1}^{n}\xi_i^V+C^G\sum_{i=1}^{n}\xi_i^G+D\sum_{i=1}^{n}\eta_i, \nonumber \\
	s.t.~~&|\langle\mathbf{w}_V,\phi_V(\mathbf{x}_i)\rangle+b_V-\langle\mathbf{w}_G,\phi_G(\mathbf{x}_i)\rangle-b_G|\leq\eta_i+\epsilon, \nonumber \\
	&y_i(\langle\mathbf{w}_V,\phi_V(\mathbf{x}_i)\rangle+b_V)\geq1-\xi_i^V, \\
	&y_i(\langle\mathbf{w}_G,\phi_G(\mathbf{x}_i)\rangle+b_G)\geq1-\xi_i^G, \nonumber \\
	&\xi_i^V\geq0, ~\xi_i^G\geq0, ~\eta_i\geq0, \nonumber \\
	&i = 1,...,n, \nonumber
\end{align}
\end{small}where the parameter $\epsilon$ constraints the closeness of the outputs of the two SVMs. $C^V$, $C^G$, and $D$ control the balance between discrimination and tolerance of noise on the training data for the two learners. In our experiments, the values of these parameters are set as $\epsilon = 0.01$, $C^V = 4$, $C^G = 4$ and $D = 0.1$. In addition, for both classifiers, we adopt radial basis function (RBF) as the kernel function.

After learning, let $\hat{\mathbf{w}}_V$, $\hat{b}_V$ represent the solution of the image features based viewpoint quality classifier and $\hat{\mathbf{w}}_G$, $\hat{b}_G$ denote the solution of geometric features based viewpoint quality classifier. Then, given a test feature vector $\mathbf{x}$ encoding the image and geometric features of a new image, its viewpoint goodness is determined by the following function
\begin{equation}
h(\mathbf{x}) = \sign(f(\mathbf{x})), 
\end{equation}
with $f(\mathbf{x})$ defined as,
\begin{equation}
\label{eq:fx}
\begin{aligned}
	f(\mathbf{x})=0.5(\langle \hat{\mathbf{w}}_V,\phi_V(x)\rangle+\hat{b}_V+\langle \hat{\mathbf{w}}_G,\phi_G(x)\rangle+\hat{b}_G).
\end{aligned}
\end{equation}


In some of the application settings, users may prefer to obtain a viewpoint goodness score instead of the binary answers output by the SVM-2K classifier. To deal with such cases, we further define a function $g(\mathbf{x})$ to measure the viewpoint goodness score by exploiting a Sigmoid function to activate the output of the linear function of SVM-2K. More specifically, we define
\begin{equation}
\begin{aligned}
	g(\mathbf{x}) = \frac{1}{1+e^{-f(\mathbf{x})}},
\end{aligned}
\end{equation}
where $f(\mathbf{x})$ is defined in Eq. \eqref{eq:fx} and the value of $g(\mathbf{x})$ is restricted to be within $[0, 1]$ with $0$ and $1$ implying that the test image is with the worst and best viewpoint, respectively.

\section{Experiments and Applications}
\label{Applications_Experiments}
\subsection{Experiments}

We perform comparative experiments on the performance of image and geometric features described in Sec.~\ref{Feature extraction}. A set of photographs taken from each of the 15 world famous architectures as well as their corresponding 3D models are used for the evaluation. Totally, we collect 5894 photos from Internet.
We also conduct an user study on the Amazon Mechanical Turk (AMT) by asking the subjects to score photo quality. A photo is assigned a score from 1 to 5. Here 1 indicates the worst viewpoint while 5 means the best. Each photo is scored 20 times by different subjects to avoid bias, with the average score defining the viewpoint goodness of a given photo. To make the training more effective, we further rule out some of the photos that have conflict scores, thus encouraging the trainer to learn from photos with unanimous scores.
At last, according to the scores, we separate the photos into two categories, and the photo with a higher and lower score is labelled as good and bad, respectively.

\subsubsection{Feature evaluation}

\begin{figure}[tb]
\centering
\includegraphics[width=1.0\linewidth]{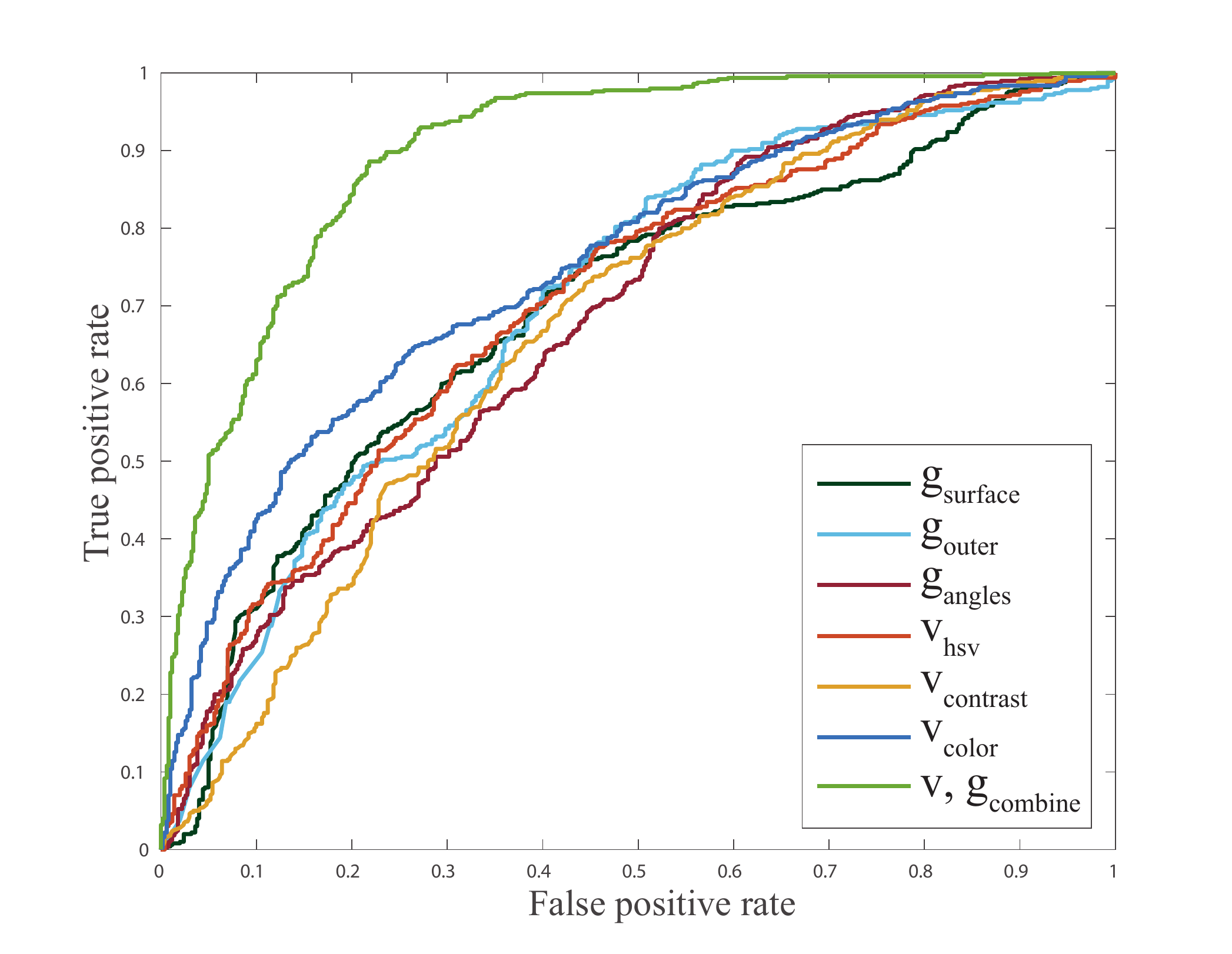}
\caption{Comparison of SVM classifier performance with different features. Combing all of the features extracted from image and geometric aspects performs better than only one of the image or geometric feature.
}
\label{fig:rocCurve}
\end{figure}

As multiple features are extracted for the learning task, we would like to first evaluate the effectiveness of each individual feature in expressing the viewpoint quality. To achieve this, we construct a SVM classifier for each feature by taking features of the gathered dataset as input and the goodness label as output. Each classifier is trained and verified with tenfold cross-validation. 
Fig.~\ref{fig:rocCurve} shows the ROC curve of top three effective image and geometric features as well as the ROC curve of the features combination involving all those described in Sec.~\ref{Feature extraction}.
As can be seen, image features of $\mathbf{v}_{color}$, $\mathbf{v}_{hsv}$, and $\mathbf{v}_{contrast}$ are effective for viewpoint selection, which suggests that the color of the photograph plays a crucial role when photographing architectures. Besides, geometric features of $\mathbf{g}_{outer}$, $\mathbf{g}_{surface}$, and $\mathbf{g}_{angles}$ achieve comparable performance with visual features, indicating that containing the whole architecture, reducing the amount of hidden surface, and selecting suitable angles have a notable impact for taking high quality pictures of architectures. At last, compared with the single feature, the combined feature outperforms all the individual features with a considerable margin, suggesting that visual and geometric features are complimentary in depicting the viewpoint goodness.

\subsubsection{Performance comparison}

\label{PerformanceComparison}

\begin{figure}[tb]
\centering
\includegraphics[width=1.0\linewidth]{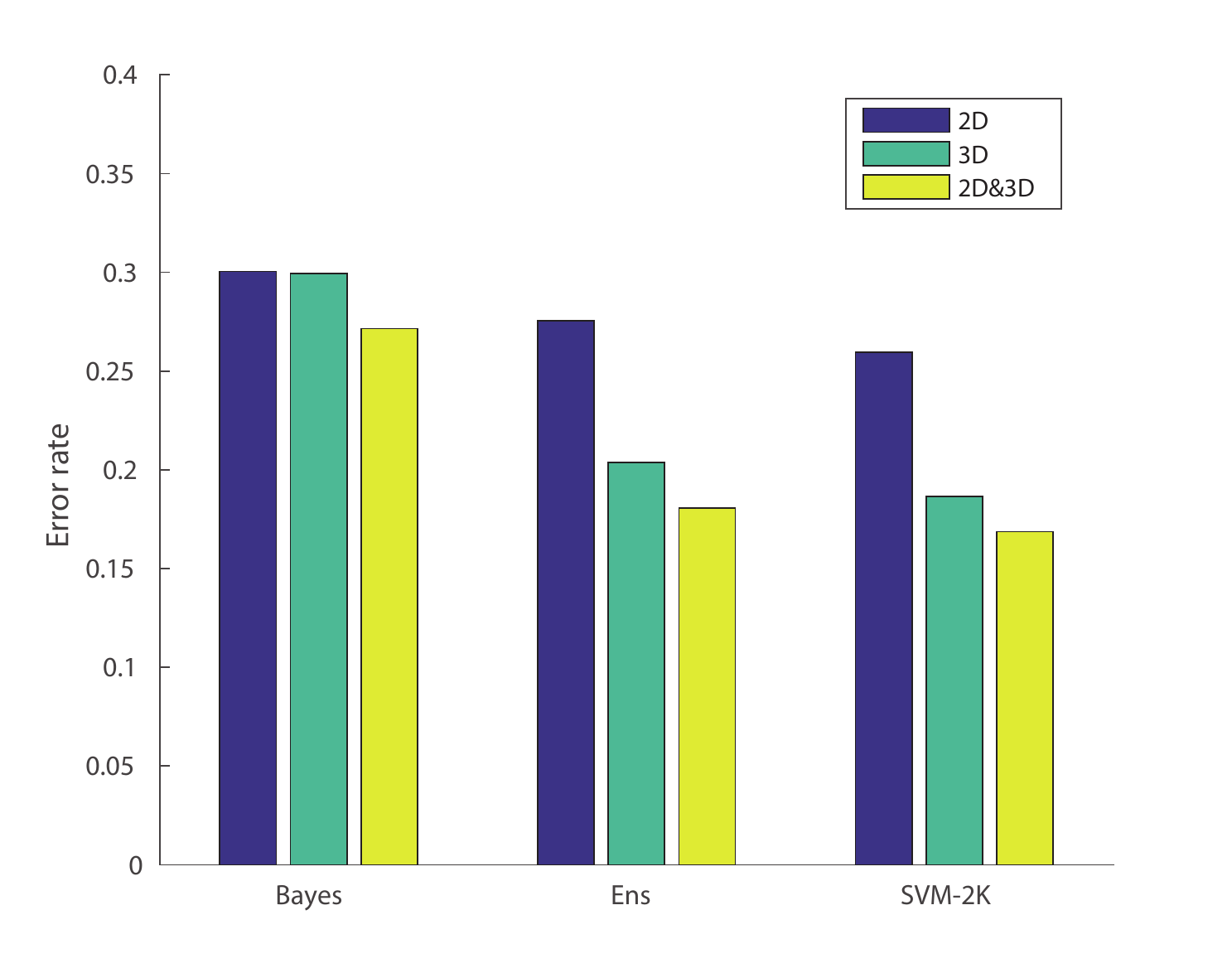}
\caption{Performance of different classifiers. As we can see, combining 2D and 3D features performs the best.}
\label{fig:combineErrRate}
\end{figure}

We choose two other different learners, namely the Bayes classifier and Ensemble classifier with random forest containing 80 trees, to train the learning models
and compare their performance with SVM-2K. All the learners are trained with tenfold cross-validation. We train these two learners on solely 2D image features or 3D geometric features as benchmarks and compare them against the learners fed with 2D-3D mixed features, which are obtained by concatenating the image and geometric feature vectors of each photo.

The overall performance comparison of the three learners is reported in Fig.~\ref{fig:combineErrRate}. Basically, all learners with 2D features achieve comparable performance while the Ensemble and SVM-2K learners obtain smaller error rates against Bayes learner when dealing with 3D and 2D-3D mixed features. SVM-2K learner outperforms Ensemble learner with a small margin. More importantly, learners trained with both image and geometric features achieve consistent superior performance over that trained with either 2D or 3D features, no matter which classifier is used.




\begin{figure*}[htb]
\centering
\includegraphics[width=0.85\paperwidth]{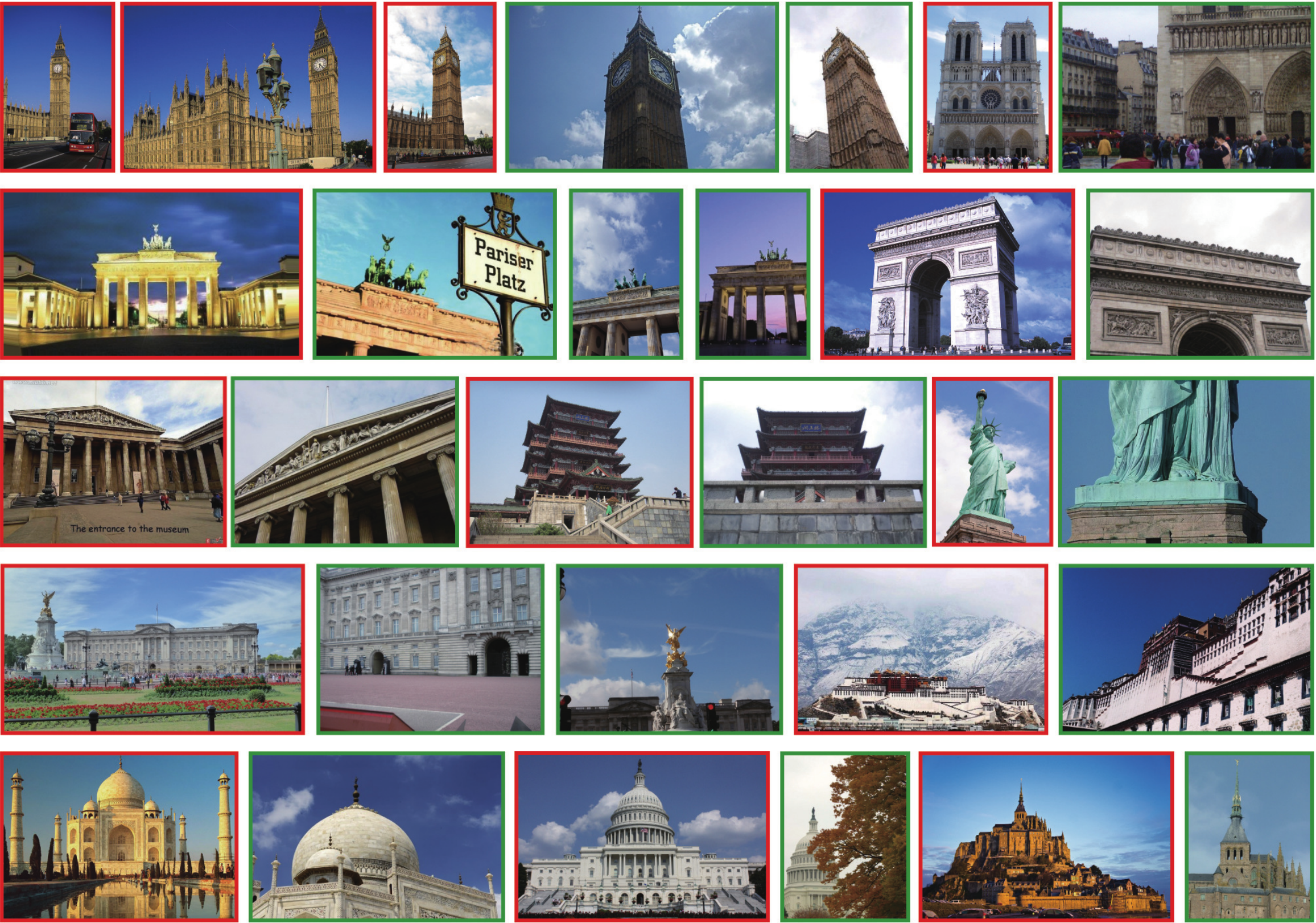}
\caption{Exemplar photos with good (marked with red frames) and bad (marked with green frames) viewpoints evaluated by our multi-view learner SVM-2K.}

\label{fig:resultForTureScene}
\end{figure*} 

We can see that using features combined with 2D and 3D will surely improve the performance of the classifier. In addition, the results also suggest that multi-view learning performs better than single-view learning for our viewpoint recommendation.
Fig. \ref{fig:resultForTureScene} presents several exemplar photos with good and bad viewpoints evaluated by our multi-view learner of SVM-2K with both 2D and 3D features, showing the effectiveness of the proposed approach.


\subsection{Applications}
Our system can assess the viewpoints effectively under the condition that both an image set and the corresponding 3D model are provided in advance. However, in real-world applications, the images and the corresponding 3D model may not be easily available at the same time. In this section, we show that our system can also be used to pick out photos with representative viewpoints from a given image set using the proposed viewpoint clustering algorithm. Thereafter, we show that our system can be flexibly adapted to deal with the scenarios where either the images or the 3D model are not available. Specifically, we demonstrate the performance of the proposed SVM-2K viewpoint learner in viewpoint recommendation for solely an image set as well as for an individual model.




\subsubsection{Representative viewpoints of world famous architectures}
Given a photo set of a world famous architecture, people may be interested in taking the pictures with representative viewpoints. In addition, finding representative viewpoints is also
especially helpful for image set navigation. Selecting images of representative viewpoints can be a natural application of the viewpoint clustering algorithm proposed in Sec.~\ref{subsec_view_clustering}. More specifically, we first estimate the model-view matrix of each input image with its corresponding 3D model and perform K-medoids clustering using the distance metric defined in Eq.~\eqref{eq:distance metric}. In the end, the photos corresponding to the medoids of all clusters are selected as those with representative viewpoints.

\begin{figure*}[!tb]
\centering
\begin{subfigure}[b]{0.48\textwidth}
\includegraphics[width=\textwidth]{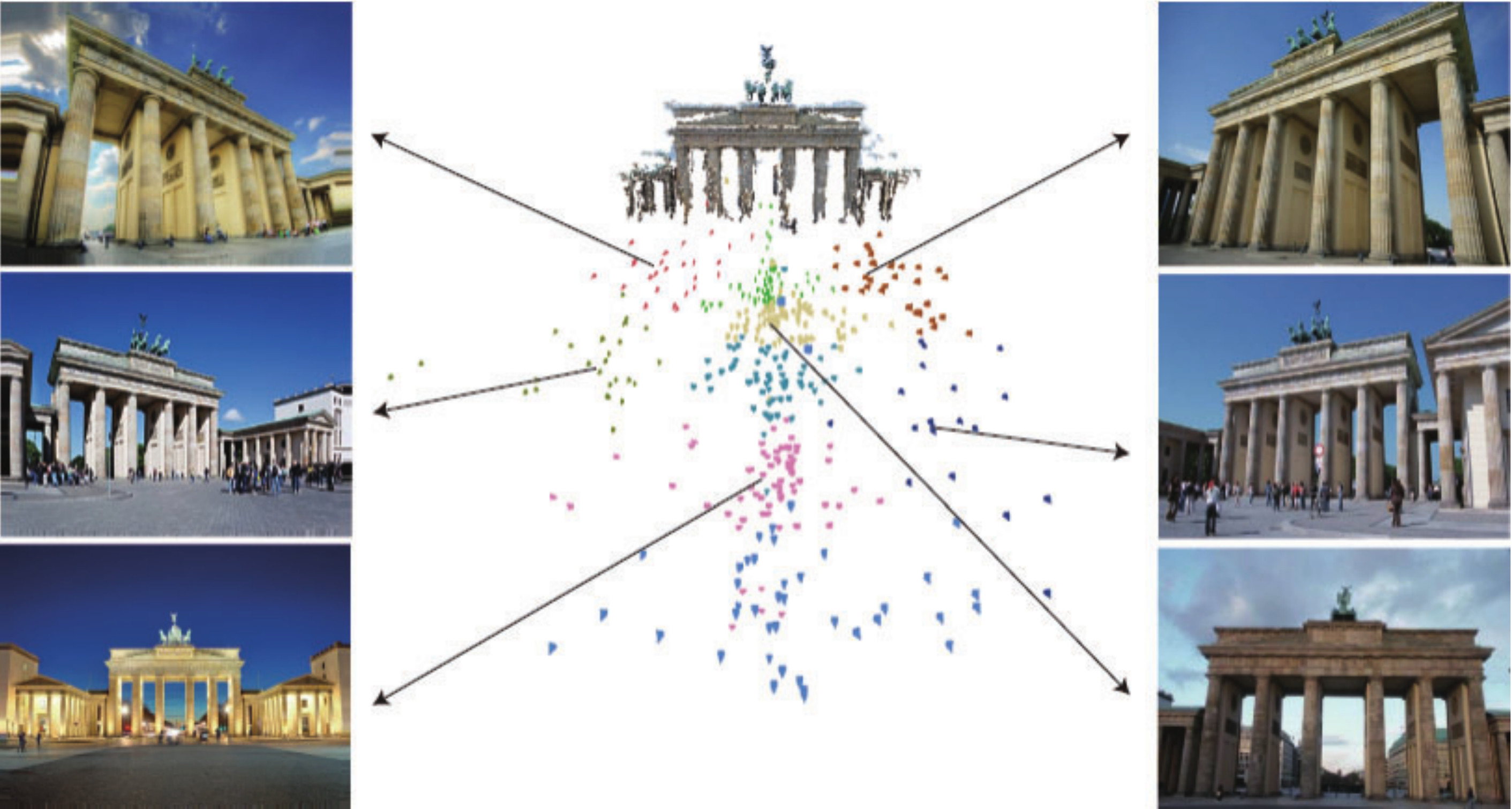} \\
\end{subfigure}
\begin{subfigure}[b]{0.48\textwidth}
\includegraphics[width=\textwidth]{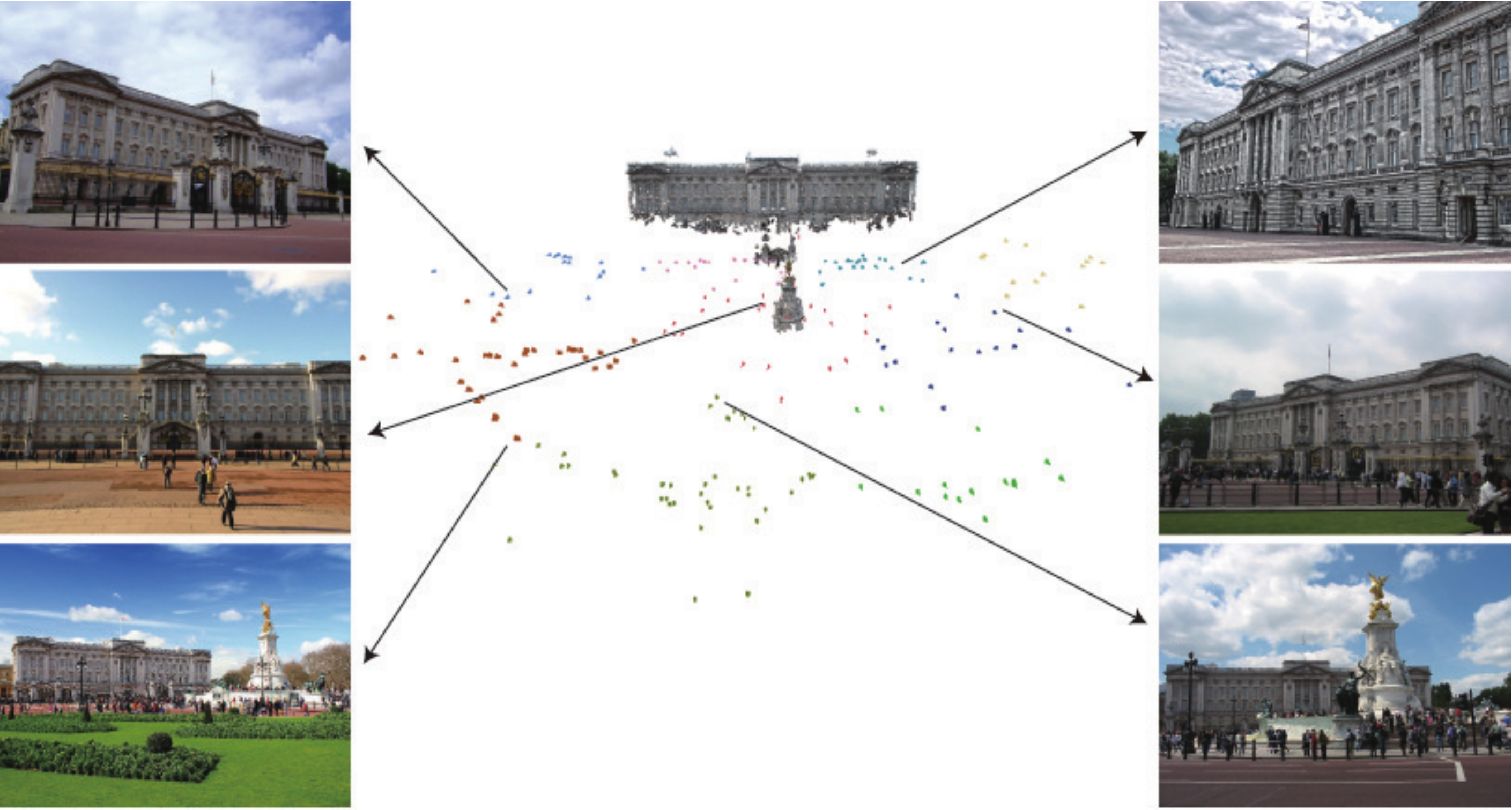} \\
\end{subfigure}

\begin{subfigure}[b]{0.48\textwidth}
\includegraphics[width=\textwidth]{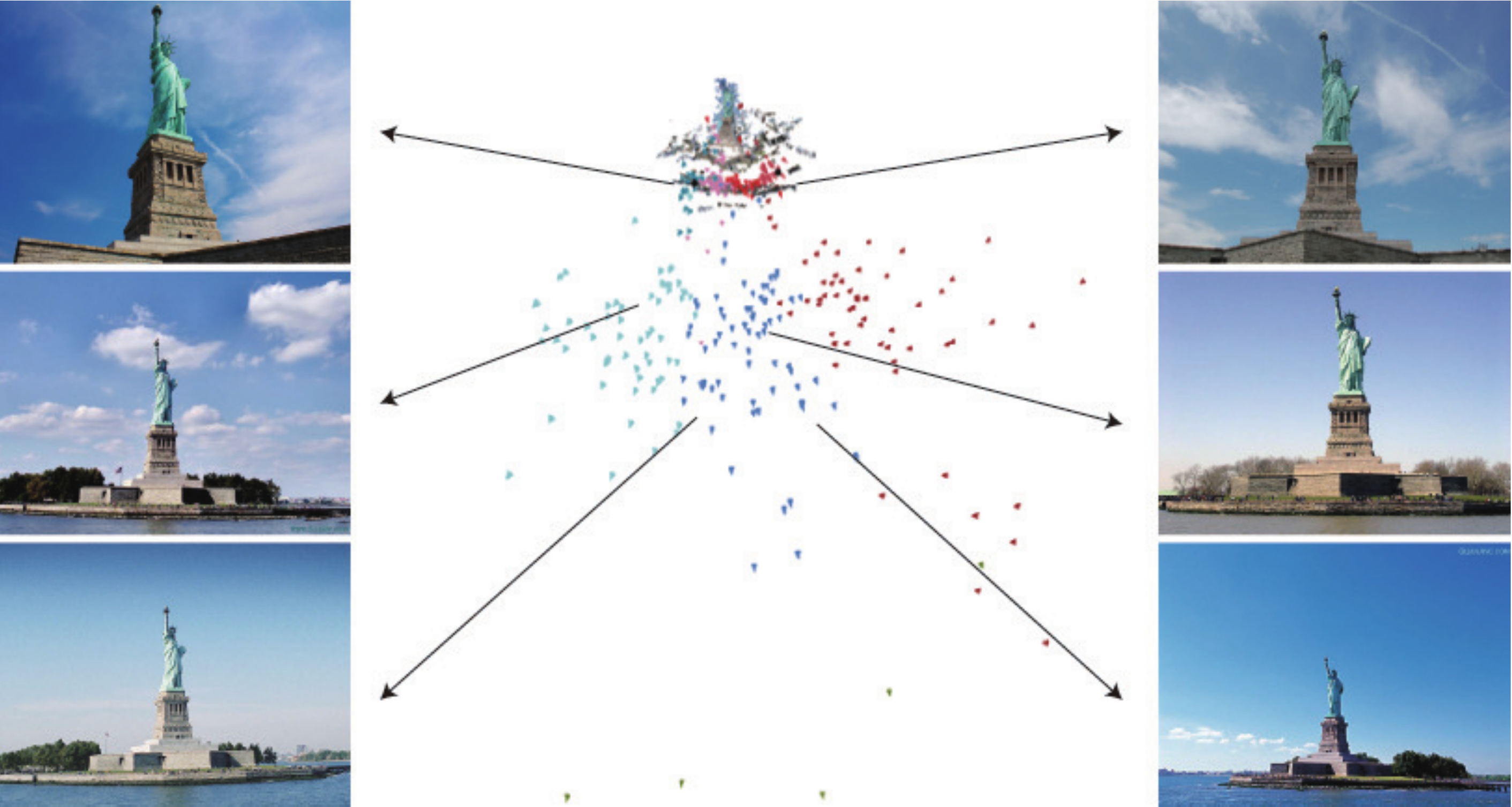}
\end{subfigure}
\begin{subfigure}[b]{0.48\textwidth}
\includegraphics[width=\textwidth]{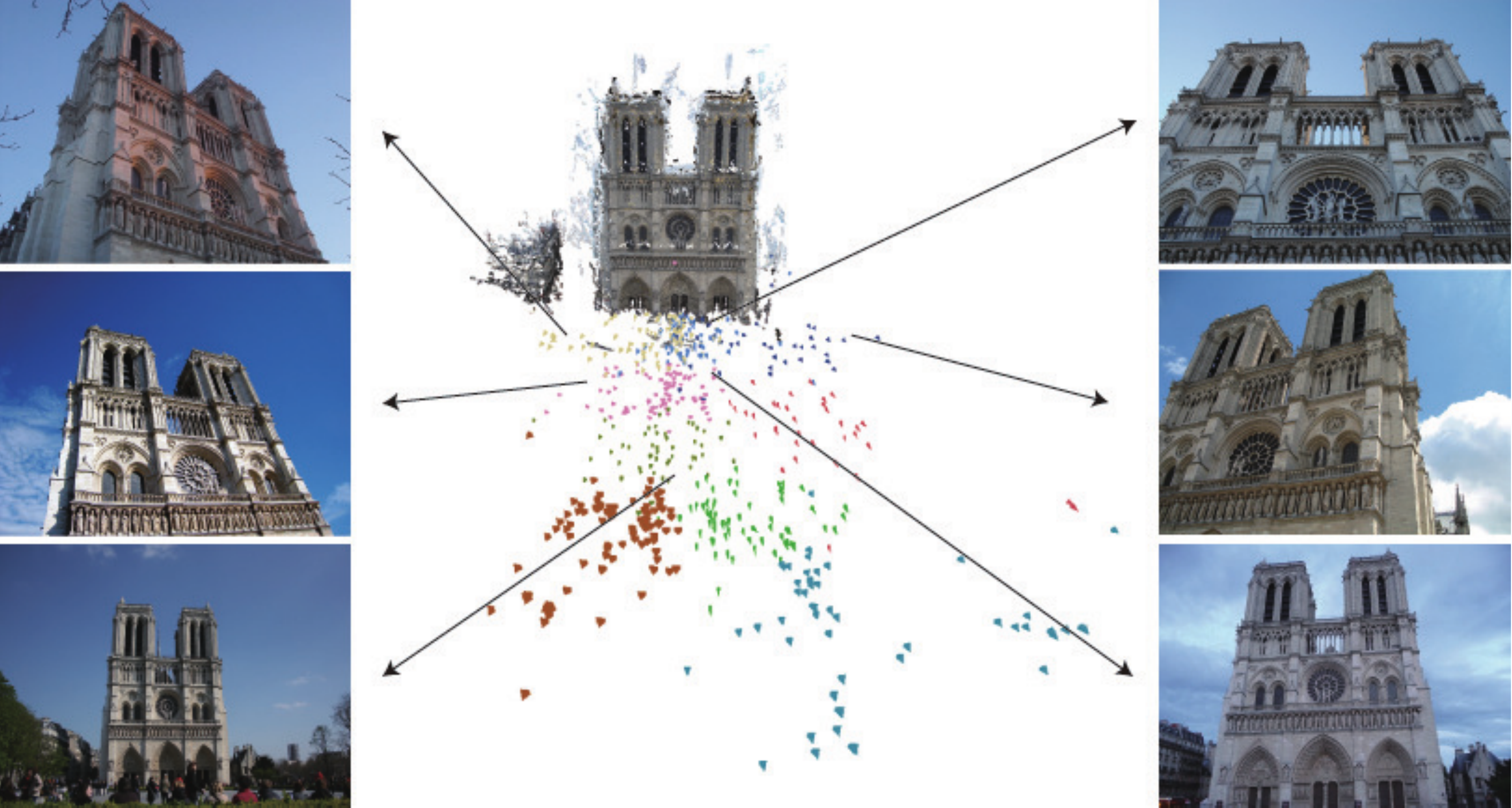}
\end{subfigure}

\caption{
Viewpoint clustering results for the Brandenburg Gate, Buckingham Palace, the Statue of Liberty, and Notre Dame cathedral with six representative viewpoints shown for each of the four landmarks.
}

\label{fig:hot places of Architecture}
\end{figure*}

In Fig.~\ref{fig:hot places of Architecture}, we show the representative photos picked out from the images we collected for four world famous architectures.
The results also reveal the  GPS locations of photographing preferred by most people when visiting these famous architectures. We will publish them to the web so that new visitors can easily choose ideal positions for taking wonderful photos.




\subsubsection{Viewpoint recommendation with only photos available}


\begin{figure}[tb]
\includegraphics[width=\linewidth]{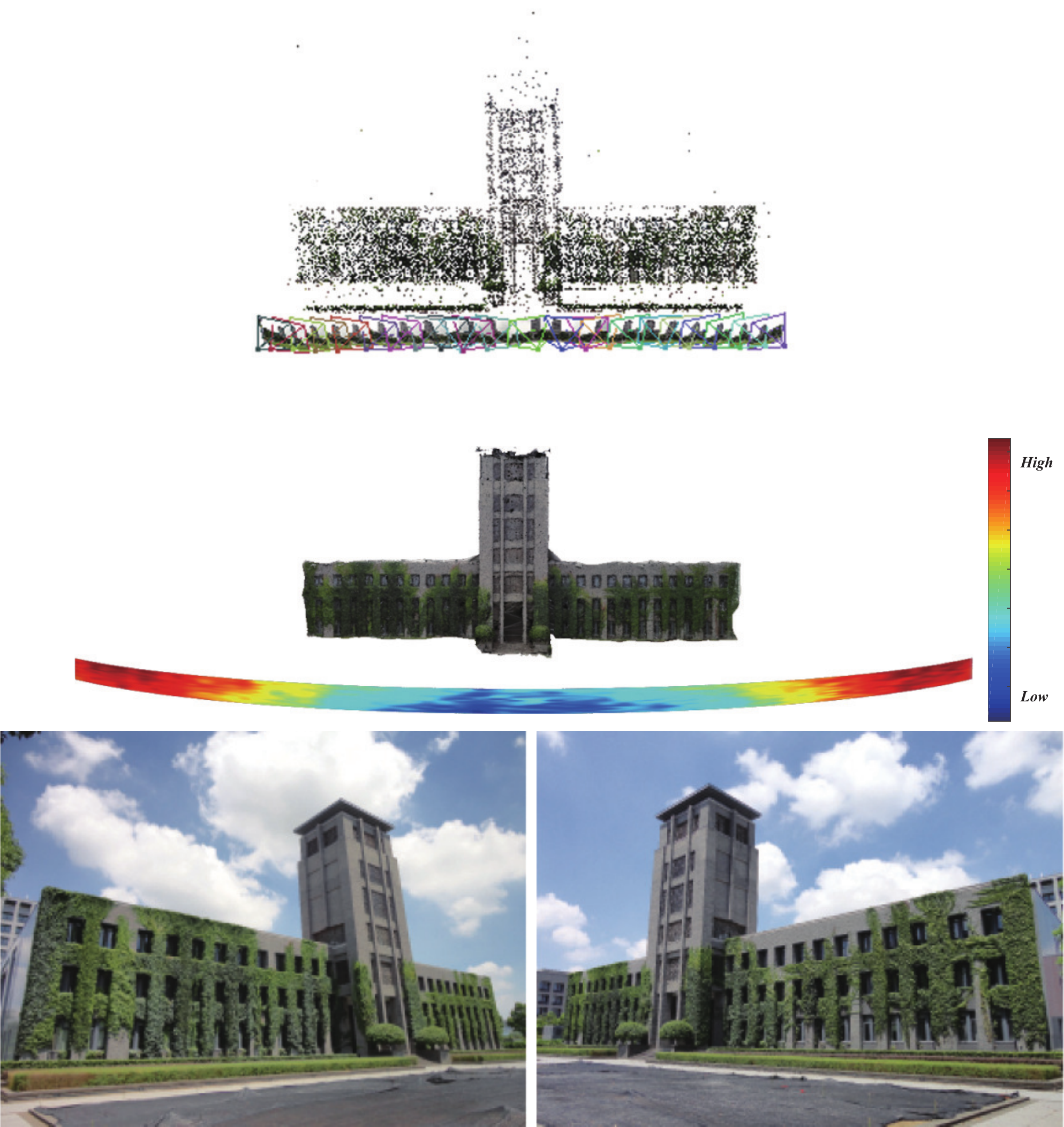}
\caption{
Top: the SfM result obtained using 23 photos for a building. Middle: visualization of the viewpoint recommendation result. Bottom: two images with the highest values of goodness of viewpoints.
}

\label{fig:appWithPhotos}
\end{figure} 

In many real-world applications, a set of pictures depicting the same architecture from different viewpoints may be provided without access to the corresponding 3D model. For example, users may take multiple photos of a building from different locations during a tour and would like to use our system to pick out the one with the best viewpoint or to help recommend the best viewpoint of the architecture of photographing. Usually, this can be achieved by training a viewpoint goodness learner based solely on image features since only input images are available. However, as we verified in Sec.~\ref{PerformanceComparison}, training from image features alone leads to a learner with relatively low performance which can be improved if 3D geometric features are incorporated.

Here, we explore the potential of the proposed multi-modular feature based SVM-2K learner in being adapted to the task of image set oriented viewpoint recommendation. The basic idea is to recover an approximate model for the architecture the input images depict and thereby extract geometric features from the model for learning. To start off with, we first generate a point cloud model by applying SfM to the input images and obtain a mesh model with Poisson reconstruction\cite{kazhdan2006poisson,Xiong:2014:RSR:2661229.2661263}. Color information is further transferred from the nodes of the point cloud model to their counterparts on the mesh model. We then extract geometric features from the reconstructed model and conduct viewpoint quality estimation for each input image using the learned SVM-2K classifier. Fig. \ref{fig:appWithPhotos} shows an example with the point cloud and mesh model constructed and two pictures of good viewpoints picked out. Note that the obtained mesh model is often incomplete since some parts of the architecture may not be contained in the input image set. This fortunately does not jeopardize the viewpoint quality estimation of the parts well reconstructed. Moreover, if sufficient input pictures are provided, the reconstructed partial mesh model can be so fine that viewpoint recommendation can be carried out on it as described in Sec. \ref{ApplicationsModel}, which is illustrated by Fig. \ref{fig:modelColorbar}.

\subsubsection{Viewpoint recommendation for a textured model}
\label{ApplicationsModel}
\begin{figure}[tb]
\centering



\input{files/fig_heatMap.tex}

\begin{figure}[tb]
\centering
\includegraphics[width=1.0\linewidth]{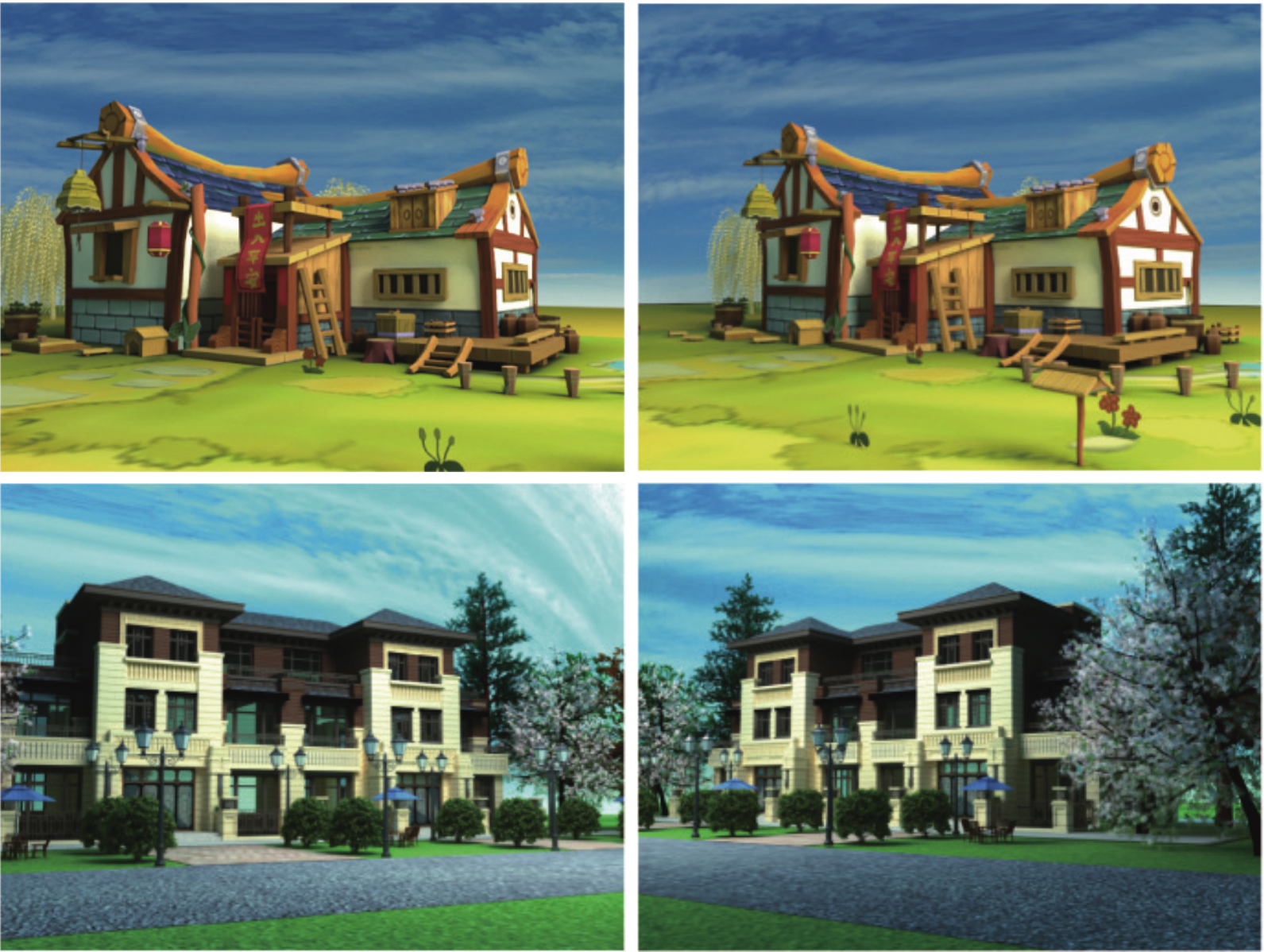}
\caption{The scenes of two other architecture models rendered under our recommended viewpoints.}
\label{fig:modelExemplars}
\end{figure}

In this subsection, we demonstrate the capacity of our framework for viewpoint goodness evaluation for a given 3D architecture model, which can be very helpful to the design and exhibition of architectures.
Specifically, we use the regression model trained with SVM-2K as described in Sec.~\ref{multi view learning}
to predict the viewpoint quality of the input model at various photographing angles.

Note that, the SVM-2K regression takes geometric features of the 3D model as well as image features of input images as input, while only a building model is provided in this case. To perform the evaluation, we uniformly sample 1024 viewpoints in a limited height range around the input model and render it at each viewpoint, obtaining 1024 images under different viewpoints. The height range starts from the horizontal line and is determined by considering the maximum reachable height level of normal photographers, covering the viewpoints people are likely to shoot at. Image and geometric features are then extracted from each rendered image and the provided 3D model, respectively. They are further fed to the SVM-2K predictor to measure the goodness of the corresponding viewpoint. After evaluating the quality of all the 1024 sampled viewpoints, the goodness of the remaining viewpoints in the reachable height range is computed by bilinear interpolation. In Fig.~\ref{fig:modelColorbar}, we visualize the prediction results of two different architecture models using the heat maps.
As shown, both models look good from the left front angle, but bad from the side angle, which plausibly agrees with visual experience. Fig.~\ref{fig:modelExemplars} shows the rendering results for the other two architecture models rendered under the viewpoints suggested by our predictor.


\section{Conclusion, Limitations and Future Work}

\label{conclusion}


We have presented a novel solution to the problem of how to choose good viewpoints for taking good photographs of architectures. We learn from photographs of world famous landmarks using both image features extracted from the images as well as geometric features computed on the 3D models under the corresponding viewpoints. Both the 2D image features and 3D geometric features are fed into SVM-2K, a multi-view learner, to learn the rules of taking high-quality photographs with good viewpoints. Our experiments show that learning from both 2D and 3D features achieves superior performance over using either of the two aspects only. The viewpoint clustering results also reveal that people do prefer some specific locations when photographing these famous landmarks.

In addition to viewpoint evaluation of photographs, our system can also be used to recommend viewpoints for photographing architectures when the user is visiting a famous tourist attraction and to suggest the viewpoints for rendering textured 3D models of buildings for the use of architectural design.

We have shown the superiority of our system but there still exist several aspects that need to be improved in the future.

\begin{itemize}

\item Since not all the handcrafted features are as effective as expected, we intend to extract more sophisticated features, such as deep features, in the future.
\item Because of the subjectivity of scores given by each individual subject in the user study,  a single averaged goodness score per photo cannot comprehensively reflect users' perception over viewpoint quality. We plan to model users' scores with some probabilistic distributions and employ Label Distribution Learning (LDL) to learn from them for better viewpoint recommendation.

\end{itemize}

\bibliographystyle{abbrv-doi}
\bibliography{paper}


\end{document}